\documentclass{article}

\usepackage{comment}
 \usepackage[main, final]{neurips_2026}
\usepackage[utf8]{inputenc} % allow utf-8 input
\usepackage[T1]{fontenc}    % use 8-bit T1 fonts
\usepackage{hyperref}       % hyperlinks
\hypersetup{hidelinks}
\usepackage{url}            % simple URL typesetting
\usepackage{booktabs}       % professional-quality tables
\usepackage{amsfonts}       % blackboard math symbols
\usepackage{nicefrac}       % compact symbols for 1/2, etc.
\usepackage{microtype}      % microtypography
\usepackage{xcolor}         % colors
\usepackage{bm}
\usepackage{bbm}
\usepackage{amsmath}
\usepackage{enumitem}
\usepackage{booktabs}
\usepackage{colortbl}
\usepackage{xcolor}
\usepackage{booktabs}
\usepackage{colortbl}
\usepackage{pifont}
\usepackage{graphicx}
\usepackage{algorithm}
\usepackage{algpseudocode}
\usepackage{multirow}
\usepackage[table]{xcolor}
\usepackage{wrapfig}
\usepackage{subcaption}
\usepackage{adjustbox}
\usepackage{tabularx}

\title{Local–Global Sparse Autoencoders for Multiscale Interpretability in Vision Models}
\author{
\textbf{Itay Benou}$^{1,2}$ \quad
\textbf{Tammy Riklin Raviv}$^{1,2}$ \\
$^{1}$The School of Electrical and Computer Engineering, Ben-Gurion University of the Negev \\
$^{2}$The School of Statistics and Data Analysis, Ben-Gurion University of the Negev \\
\texttt{benoui@post.bgu.ac.il}, \texttt{rrtammy@bgu.ac.il}
}

\begin{document}

\maketitle

% \begingroup
% \renewcommand{\thefootnote}{}
% \footnotetext{Project code: \url{https://https://github.com/itaybenou/LG-SAE}}
% \endgroup

\begin{abstract}
Visual understanding is inherently hierarchical, as high-level semantic concepts are composed of fine-grained, spatially localized components. Yet existing interpretability methods typically focus on a single feature scale, either local or global, leaving a critical gap in understanding the compositional structure between them. Recovering this structure without supervision is particularly challenging, as the correspondence between local and global concepts is latent, varies across samples, and may emerge only through inconsistent or partial patterns. To address this challenge, we introduce Local--Global Sparse Autoencoders (LG-SAE), a multiscale interpretability framework for vision models. The core innovation of LG-SAE is a learned bridge that explicitly captures the compositional relationships between local and global visual concepts. By jointly operating on patch-level and image-level embeddings, LG-SAE learns a sparse, interpretable graph that reveals how high-level semantic concepts are constructed from a small set of spatially grounded components. Building on this representation, we present a unified framework for discovering local-global concept relationships, explaining model predictions through multiscale Concept Bottleneck Models, and steering models via localized concept edits. Extensive evaluations show that our framework: (i) captures meaningful cross-scale concept relations; (ii) enables high-quality spatial grounding without compromising downstream accuracy; (iii) provides human-interpretable model explanations, as validated by a user study; and (iv) facilitates intuitive and effective model steering.
Project code: \url{https://github.com/itaybenou/LG-SAE}.
\end{abstract}

% #################################################
%                   Introduction
% #################################################

\section{Introduction}

\begin{figure}[t]
    \centering
    \includegraphics[width=\linewidth]{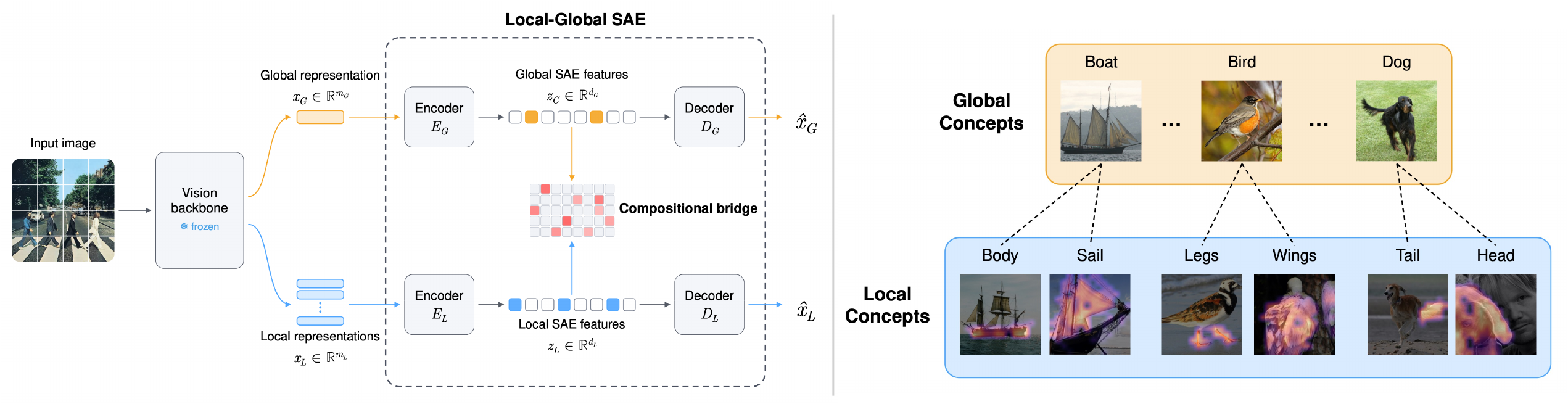}
    \caption{\textbf{Local--Global SAE.} Left: By jointly operating on global and local model representations, LG-SAE learns a sparse compositional bridge between their SAE features. Right: The learned compositional graph reveals meaningful structural connections between local and global concepts.}
    \label{fig:LG_SAE_overview}
\end{figure}

Visual understanding is inherently hierarchical, as high-level semantic concepts are composed of simpler, spatially localized elements. Consider, for example, an image of a street classified by a vision model, as shown in Fig.~\ref{fig:LG_SAE_overview}. The prediction may rely on visual cues such as pedestrians, crosswalk, road, and cars. While some cues are localized, others recur across image regions; together, they support a higher-level semantic concept that is not tied to any single location. Understanding such a prediction therefore requires more than identifying image-level concepts~\citep{kim2018interpretability, ghorbani2019towards, zhang2021invertible, fel2023craft} or highlighting salient image regions~\citep{selvaraju2017grad, srinivas2019full, zhou2016learning, binder2016layer, sundararajan2017axiomatic, chefer2021transformer}. It requires explaining the compositional structure that connects these two levels: how spatially localized visual evidence is assembled into a global semantic representation.

Modern vision neural networks naturally encode information at multiple scales. In Vision Transformers (ViTs)~\citep{dosovitskiy2020image}, for example, patch tokens retain spatially localized information, while the \texttt{CLS} token aggregates them into an image-level representation. However, existing interpretability methods rarely model the relationships between these scales. Sparse Autoencoders (SAEs)~\citep{lim2024sparse, fel2025archetypal, thasarathan2025universal, fel2025into, stevens2025sparse, pach2025sparse, kulkarni2025interpretable} and Concept Bottleneck Models (CBMs)~\citep{koh2020concept, yuksekgonul2022post, yang2023language, panousis2023sparse, oikarinen2023label, rao2024discover, prasse2024dcbm, srivastava2024vlg} have been widely used to interpret vision models, but are generally applied to either image-level or patch-level representations. As a result, they may identify a global concept such as ``street'', or that the image contains a car, yet they are unable to explain how local elements combine to form that global concept.

This missing multiscale structure limits the explanatory power of current interpretability methods. Recent SAE variants such as Matryoshka SAEs~\citep{zaigrajew2025interpreting, bussmann2025learning} and Matching Pursuit SAEs~\citep{costa2025flat} capture hierarchical relations among learned features, while CBM approaches such as SALF-CBM~\citep{benou2025show} and CFM~\citep{wittenmayer2026cfm} improve the spatial grounding of concepts. However, both lines of work largely remain single-scale and do not explicitly capture compositional relations between local and global concepts.

To address this gap, we introduce \textbf{Local--Global Sparse Autoencoders (LG-SAE)}, a multiscale interpretability framework for vision models. By jointly operating on patch-level and image-level representations, LG-SAE learns a sparse compositional graph that links global semantic concepts to their spatially grounded visual components. Our main contributions are:
\begin{itemize}[leftmargin=*]
    \item \textbf{LG-SAE: Local--Global Structure Discovery.}
    We jointly discover local and global concepts and explicitly learn the compositional links between them.

    \item \textbf{LG-CBM: Multiscale Concept Bottleneck Model.}
    We explain model predictions through multiscale, spatially grounded concepts. To provide human-interpretable explanations, we introduce a spatially aware concept naming method based on CLIP-Dissect~\citep{oikarinen2022clip}.

    \item \textbf{Multiscale Model Steering.}
    We extend CBM interventions beyond global concepts by enabling localized edits to spatial concept activations, allowing targeted steering of model predictions.

    \item \textbf{Empirical Validation.}
    Experiments on COCO-Stuff, Part-ImageNet, ImageNet, and Places validate compositional structure recovery, spatial concept grounding while maintaining competitive downstream performance, and effective model steering. We further validate interpretability and concept naming via a user study.
\end{itemize}

% #################################################
%                   BACKGROUND
% #################################################

\section{Background: Sparse Autoencoders}
\label{background}

Sparse autoencoders (SAEs) decompose model representations $x \in \mathbb{R}^m$ into an overcomplete sparse code $z \in \mathbb{R}^d$ by reconstructing the input as a linear combination of a learned dictionary $D$:
\begin{equation} \label{eq:saes}
    {z} = {\sigma}\!\left({W}^{\top} {x} + {b}_{\text{enc}}\right), \quad
    \hat{{x}} = {D} {z} + {b}_{\text{dec}},
\end{equation}
where $\{W, b_{\text{enc}}\}$ and $\{D, b_{\text{dec}}\}$ represent the parameters of the encoder and decoder, respectively, and \({\sigma}\) denotes a nonlinearity, e.g., ReLU~\citep{bricken2023towards,cunningham2023sparse}, TopK~\citep{makhzani2013k,gao2024scaling}, BatchTopK~\citep{bussmann2024batchtopk}, and JumpReLU~\citep{rajamanoharan2024jumping,rajamanoharan2024improving}.
The model is trained by minimizing a reconstruction objective:
\begin{equation} \label{eq:sae_loss}
\mathcal{L}_{\text{sae}} = \left\lVert {x} - \hat{{x}} \right\rVert_2^2 + \lambda\, \mathcal{R}({z}) + \alpha_{aux}\, \mathcal{L}_{\text{aux}},
\end{equation}
where \(\mathcal{R}({z})\) is a regularization term for promoting sparsity of SAE features $z$, and \(\mathcal{L}_{\text{aux}}\) is an auxiliary term used to prevent dead codes in ${z}$~\citep{gao2024scaling}.

% #################################################
%                   LG-SAE
% #################################################

\section{Local--Global Sparse Autoencoders}
\label{lg_sae}

We introduce Local--Global Sparse Autoencoders (LG-SAE), extending SAEs to jointly decompose local and global model representations.
Formally, given a pretrained vision model, we consider a global (image-level) representation $x_\mathrm{G} \in \mathbb{R}^{m_\mathrm{G}}$ and a local (dense) representation $X_\mathrm{L} \in \mathbb{R}^{P \times m_\mathrm{L}}$ 
% $x_L \in \mathbb{R}^{n \times m_L}$
within the model, where $P$ denotes the spatial dimension and ${x}_{\mathrm{L},p}$ denotes the local representation at spatial location $p \in \{1, \ldots, P\}$. In vision transformers (ViTs), these can represent the \texttt{CLS} and patch tokens of a specific layer, or across different layers.
Since representations at different scales may have different distributions or dimensionalities, we decompose $x_\mathrm{G}$ and ${x}_{\mathrm{L},p}$ into sparse, interpretable features using separate global and local SAEs, each reconstructing its corresponding input:

% We introduce Local--Global Sparse Autoencoders (LG-SAE), extending SAEs to jointly decompose local and global model representations.
% Given a pretrained vision model, we consider a global representation $x_G \in \mathbb{R}^{m_G}$ and a local representation $X_\text{L} \in \mathbb{R}^{P \times m_L}$, where $P$ is the spatial dimension and ${x}_{\text{L},p}$ denotes the representation at location $p$.
% In vision transformers (ViTs), these may correspond to \texttt{CLS} and patch tokens from the same or different layers.
% Because the two scales may differ in distribution and dimensionality, we use separate global and local SAEs to decompose $x_G$ and ${x}_{\text{L},p}$ into sparse, interpretable features while reconstructing their respective inputs:

\begin{equation} \label{eq:lg_saes}
\begin{aligned}
    {z}_{\text{G}} &= {\sigma}\!\left({W}_{\text{G}}^{\top} {x}_{\text{G}} + {b}_{\text{G,enc}}\right), \quad 
    \hat{{x}}_{\text{G}} = {D}_{\text{G}} {z}_{\text{G}} + {b}_{\text{G,dec}}, \\
    {z}_{\text{L},p} &= {\sigma}\!\left({W}_{\text{L}}^{\top} {x}_{\text{L},p} + {b}_{\text{L,enc}}\right), \quad 
    \hat{{x}}_{\text{L},p} = {D}_{\text{L}} {z}_{\text{L},p} + {b}_{\text{L,dec}},
\end{aligned}
\end{equation}

where ${z}_{\text{G}} \in \mathbb{R}^{d_{\text{G}}}$ and ${z}_{\text{L},p} \in \mathbb{R}^{d_{\text{L}}}$ denote the global and local SAE features, respectively. Since the global SAE reconstructs image-level representations, its features tend to capture high-level semantics, whereas the local SAE captures finer-grained, spatially localized components. We allow different global and local dictionary sizes $d_{\mathrm{G}}$ and $d_{\mathrm{L}}$ to reflect scale-specific feature complexity.

A key component of LG-SAE is a sparse bridge matrix
$\mathcal{C} \in \mathbb{R}^{d_{\mathrm{G}} \times d_{\mathrm{L}}}$,
which models the compositional relationship between global and local SAE features.
Given corresponding global and local representations, the goal of the bridge is to approximate the global
activations $z_{\mathrm{G}}$ as a sparse combination of spatially pooled local
activations. Formally, let
$\bar{z}_{\mathrm{L}}=\phi(\{z_{\mathrm{L},p}\}_{p=1}^{P})
\in \mathbb{R}^{d_{\mathrm{L}}}$
denote the pooled local activations, where $\phi$ is a spatial pooling operator.
We define
\begin{equation}
\label{eq:C_bridge}
z_{\mathrm{G},\mathrm{app}} =
\mathrm{ReLU}\!\left(\mathcal{C}\bar{z}_{\mathrm{L}} + b\right),
\end{equation}
where $b$ is a bias term.
The bridge can therefore be interpreted as a cross-scale graph, where
$\mathcal{C}_{i,j}$ measures the contribution of local feature $j$ to the
approximation of global feature $i$.
% To keep this graph interpretable, we enforce sparsity on $\mathcal{C}$ such that each global concept is composed of a small subset of local features.
To keep this graph interpretable, we promote sparsity in $\mathcal{C}$ and enforce non-negativity by clamping its entries whenever the bridge is used, such that each global concept is composed of a small subset of local features.
By integrating $\mathcal{C}$ directly into training, as described below, LG-SAE induces a sparse compositional alignment between independently learned local and global SAE features.

\subsection{Training Process}
\label{training_process}

Training LG-SAE involves optimizing the global and local SAEs along with the compositional bridge. Joint optimization of these components can create unstable ``moving targets'', as the bridge simultaneously attempts to align evolving local and global features. To avoid unstable co-adaptation, we adopt a three-stage training strategy.
% To stabilize optimization and preserve the interpretability of the learned features, we adopt a three-stage training strategy that separates multiscale concept discovery from compositional alignment.

\textbf{1. Multiscale concept discovery.}
We first train the global and local SAEs independently according to Eq.~\ref{eq:sae_loss}. This stage yields stable, scale-specific sparse feature spaces, with global features $z_{\mathrm{G}}$ capturing image-level semantics and local features $z_{\mathrm{L}}$ capturing spatially localized visual components.

\textbf{2. Learning compositional relations.}
We next freeze the parameters of both SAEs and learn the compositional relations between their features by optimizing only the bridge parameters $\mathcal{C}$ and $b$. The bridge is trained using two complementary objectives: an alignment loss that encourages the global approximation derived from local features to match the global SAE activations, and a reconstruction loss that ensures this approximation retains sufficient information to reconstruct the original global representation through the frozen global SAE decoder, i.e.,
\begin{equation}
\label{eq:link_loss}
\mathcal{L}_{\mathrm{align}}
=
\|z_{\mathrm{G}} - z_{\mathrm{G},\mathrm{app}}\|_2^2,
\qquad
\mathcal{L}_{\mathrm{rec}}
=
\|x_{\mathrm{G}} -
(D_{\mathrm{G}}z_{\mathrm{G},\mathrm{app}} + b_{\mathrm{G,dec}})\|_2^2.
\end{equation}
% Second, a reconstruction loss encourages the global approximation to preserve sufficient information to reconstruct the original global model representation through the frozen global SAE decoder:
% \begin{equation}
% \label{eq:fidelity_loss}
% \mathcal{L}_{\mathrm{rec}}
% =
% \|x_{\mathrm{G}} -
% (D_{\mathrm{G}}z_{\mathrm{G},\mathrm{app}} + b_{\mathrm{G,dec}})\|_2^2.
% \end{equation}
The total bridge objective is
\begin{equation}
\label{eq:bridge_loss}
\mathcal{L}_{\mathrm{bridge}}
=
\mathcal{L}_{\mathrm{align}}
+
\alpha \mathcal{L}_{\mathrm{rec}}
+
\lambda_{\mathcal{C}}\mathcal{R}(\mathcal{C}),
\end{equation}
where $\mathcal{R}(\mathcal{C})$ is a regularization term that promotes sparsity in $\mathcal{C}$, e.g., an $L_1$ penalty or a Top-K constraint.

\textbf{3. Iterative refinement.}
The first two stages establish a sparse mapping between independently learned local and global features.
To encourage the local SAE features to support the recovered compositional structure, we introduce an iterative co-adaptation stage that alternates between two optimization steps until convergence:

\begin{itemize}[leftmargin=18pt]
    \item \textbf{Local SAE refinement.}
    With the global SAE and bridge parameters fixed, we update the local SAE using a composition-aware objective:
    \begin{equation}
    \label{eq:refine_loss}
    \mathcal{L}_{\mathrm{refine}}
    =
    \mathcal{L}_{\mathrm{sae}}
    +
    \beta
    \|z_{\mathrm{G}} - z_{\mathrm{G},\mathrm{app}}\|_2^2.
    \end{equation}
    The SAE objective preserves the reconstruction quality and sparsity of local features, while the additional term encourages them to better align with the learned cross-scale relationships.

    \item \textbf{Bridge refinement.}
    We freeze the refined local SAE and update the bridge parameters $\mathcal{C}$ and $b$ using the $\mathcal{L}_{\mathrm{bridge}}$ objective from Eq.~\ref{eq:bridge_loss}, allowing the bridge to adapt to the updated local features.
\end{itemize}

% #################################################
%                   LG-CBM
% #################################################

\section{Multiscale Interpretability of Vision Models}
\label{explainability_framework}
% 
% Say something on how our goal is to make scale-aware interpretability easy, efficient and accessible for any vision models.
% Building on top of LG-SAE, we present a unified, end-to-end framework which transforms any pre-trained vision backbone into an inherently explainable model. 
% Building on the proposed LG-SAE, we present a unified, end-to-end framework for interpretability and steering of pre-trained vision models. 
% In the sections below we describe how this framework facilitates: 
% (1)~\textit{Local--Global Concept Bottleneck Models (LG-CBM)}, transforming opaque backbones into interpretable models with spatially grounded concept-based explanations; 
% (2)~\textit{Spatially Aware Concept Naming}, assigning human-interpretable labels to discovered features; and 
% (3)~\textit{Multiscale Model Steering}, enabling users to adjust predictions through targeted edits to local concept activations.

Building on LG-SAE, we introduce a unified, end-to-end framework that enables: (1)~\textit{Local--Global Concept Bottleneck Models} (LG-CBM), which provide multiscale, spatially grounded concept-based explanations with spatially aware concept naming; and (2)~\textit{Multiscale Model Steering}, which enables user intervention through targeted modifications of local concept activations.

% spatially-grounded concept bottleneck model. XXX, concept discovery and compositionality learning are completely unsupervised. Our framework: (1)~Provides human-interpretable explanations that are visually grounded in the input image; (2)~XXX automatic and spatially-aware concept naming; (3)~Enables local model steering via XXX user interventions.

% To be added. Explain how by simple training a classification head on top of XXX-SAE, the method can be used to transform any pre-trained vision model into an explainable XXX-CBM that allows: (1) Spatially-grounded concept-based explanations; (2) Concept discovery and hierarchy learning are completely unsupervised; (3) At inference, only the global SAE (lighter) is served as a bottleneck layer, and only activated concepts are grounded using the local SAE, saving significant memory and compute; (4) Allows interactive local interventions. 

\subsection{Local--Global Concept Bottleneck Models}
\label{LG-CBMs}

CBMs explain predictions by projecting model representations into human-interpretable concepts and learning a downstream predictor over their activations. However, they typically operate at a single scale, providing either image-level explanations or spatial concept maps. Here, we leverage LG-SAE to construct a multiscale \emph{Local--Global Concept Bottleneck Model} (LG-CBM) that provides concept-based explanations that are semantically meaningful, compositional, and spatially grounded.

% Concept Bottleneck Models (CBMs) explain predictions through human-interpretable concept activations, but typically operate at a single scale. We use LG-SAE to construct a \emph{Local--Global Concept Bottleneck Model} (LG-CBM) that links global prediction concepts to localized visual evidence.

\textbf{Training.}
Let $f$ be a pretrained backbone and $\{(I_n,y_n)\}_{n=1}^{N}$ a labeled dataset of $N$ images. For each image, we extract a global representation $x_{\mathrm{G}}$, e.g., the \texttt{CLS} token, and local representations $X_{\mathrm{L}}$, e.g., patch tokens. We train LG-SAE as described in Section~\ref{training_process}, yielding global concept activations $z_{\mathrm{G}}$ and local concept maps $Z_{\mathrm{L}}=\{z_{\mathrm{L},p}\}_{p=1}^{P}$. The compositional relationships between them are captured by the sparse bridge matrix $\mathcal{C}$.
For downstream classification, we use $z_{\mathrm{G}}$ as the concept bottleneck and train a linear classification head, i.e., $\ell(I)=W_{\mathrm{cls}}z_{\mathrm{G}}+b_{\mathrm{cls}},$
% \begin{equation}
% \ell(I)=W_{\mathrm{cls}}z_{\mathrm{G}}+b_{\mathrm{cls}},
% \end{equation}
where $W_{\mathrm{cls}}\in\mathbb{R}^{C\times d_{\mathrm{G}}}$ and $b_{\mathrm{cls}}\in\mathbb{R}^{C}$ are learned with a standard cross-entropy loss. The predicted class is given by $\hat{y}=\arg\max_c \ell_c(I)$.

\textbf{Multiscale explainability.}
Unlike standard CBMs, LG-CBM provides compositional model explanations at inference time by grounding global concepts in localized visual evidence.
\\
\textit{(i)~Global explanations:}
Given a model prediction $\hat{y}$ for a test image, we define the contribution of global concept $i$ to the predicted class as:
\begin{equation} 
\label{eq:contribution_score}
\mathrm{Contrib}_i(I) = W_{\mathrm{cls}}[\hat{y}, i] \cdot z_{\mathrm{G}}[i].
\end{equation}
Since $z_{\mathrm{G}}$ is sparse, the model's prediction is highly interpretable and can be explained by a small set of global concepts, based on their contribution scores.
\\
\textit{(ii)~Local explanations:}
We provide spatially grounded evidence for an activated global concept~$i$. 
% we leverage the learned bridge matrix $\mathcal{C}$.
Given the pooled local activations $\bar{z}_{\mathrm{L}} = \phi(Z_{\mathrm{L}})$, we define the relevance of each local concept $j$ to global concept $i$ based on its activation strength and its connection through the bridge:
\begin{equation}
s_{i,j}(I) = \mathcal{C}[i,j] \cdot \bar{z}_{\mathrm{L}}[j].
\end{equation}
Since both $\mathcal{C}$ and $\bar{z}_{\mathrm{L}}$ are sparse, the activation of global concept~$i$ can be visually grounded by a small set of its most relevant local constituents, using their corresponding spatial activation maps $Z_{\mathrm{L}}[:,j]$.

\subsection{Spatially Aware Concept Naming}
\label{concept_naming}
CBMs commonly assign human-interpretable semantic labels to concepts~\citep{oikarinen2023label, rao2024discover}. Methods such as CLIP-Dissect~\citep{oikarinen2022clip} typically match concept activations with image-level descriptions from vision-language models. 
However, this global approach can produce spurious labels for spatially localized features. For example, a feature activated by a license plate may be labeled as ``car'' due to the surrounding context. To address this limitation, we introduce a spatially aware extension of CLIP-Dissect that matches local concept activations with dense patch-level similarities from CLIP-DINOiser~\citep{wysoczanska2024clip}, which provides spatially aligned CLIP embeddings.

Formally, let $\mathcal{D}_{\mathrm{probe}}=\{I_n\}_{n=1}^{N}$ be a probe dataset and $\mathcal{T}=\{t_m\}_{m=1}^{M}$ a vocabulary of candidate labels. For each image $I_n$, CLIP-DINOiser produces a visual embedding $u_{n,p} \in \mathbb{R}^{d}$ at each spatial location $p$, while its text encoder maps each label $t_m$ to an embedding $v_m \in \mathbb{R}^{d}$. We compute spatial image--text similarities
$S[n,p,m] = \cos(u_{n,p}, v_m)$,
which measure the correspondence between candidate label $t_m$ and spatial location $p$ in image $I_n$.
Let $\mathcal{Z}\in\mathbb{R}^{N\times P\times d_{\mathrm{L}}}$ denote the local SAE activations over the entire probe set. We measure the association between local feature $j$ and candidate label $m$ using patch-level Soft-WPMI~\citep{oikarinen2022clip}:
\begin{equation}
\alpha(m,j) =
\mathrm{SoftWPMI}\!\left(S_{m}, \mathcal{Z}_{j}\right),
\end{equation}
where $S_{m}=S[:,:,m]$ and $\mathcal{Z}_{j}=\mathcal{Z}[:,:,j]$ are defined over the same image--location grid $(n,p)$.
We assign each local feature $j$ the label with the highest association score,
$m^*(j)=\arg\max_m \alpha(m,j)$.

% Let $\mathcal{D}_{\mathrm{probe}}=\{I_n\}_{n=1}^{N}$ be a probe dataset and $\mathcal{T}=\{t_m\}_{m=1}^{M}$ a candidate-label vocabulary. For each image $I_n$, CLIP-DINOiser produces a visual embedding $u_{n,p}\in\mathbb{R}^d$ at spatial location $p$, while its text encoder maps label $t_m$ to $v_m\in\mathbb{R}^d$. We compute
% \[
% S[n,p,m]=\cos(u_{n,p},v_m).
% \]
% Let $\mathcal{Z}\in\mathbb{R}^{N\times P\times d_{\mathrm{L}}}$ denote the local SAE activations over the probe set. We measure the association between local feature $j$ and label $m$ using patch-level Soft-WPMI~\citep{oikarinen2022clip}:
% \begin{equation}
% \alpha(m,j)=
% \mathrm{SoftWPMI}\!\left(S_m,\mathcal{Z}_j\right),
% \end{equation}
% where $S_m=S[:,:,m]$ and $\mathcal{Z}_j=\mathcal{Z}[:,:,j]$ are defined over the same image--location grid. We assign feature $j$ the label
% $m^*(j)=\arg\max_m \alpha(m,j)$.

\subsection{Multiscale Model Steering}
\label{steerability}
A major advantage of CBMs is their steerability, which enables interventions on concept activations to assess their effects on model predictions. Beyond standard CBM interventions on global concepts, LG-CBM supports multiscale model steering through \textit{localized interventions} that modify local concept activations within specific spatial regions.
Given a local concept $j$ and an edit region specified by a binary mask $M \in \{0,1\}^P$, we define a localized intervention as
\begin{equation}
Z'_{\mathrm{L}}[p,j]
=
\begin{cases}
(1+\delta M[p]) Z_{\mathrm{L}}[p,j], & \delta \in [-1,0] \\[4pt]
Z_{\mathrm{L}}[p,j] + \delta M[p], & \delta > 0,
\end{cases}
\end{equation}
where $\delta$ controls the intervention strength: values of $\delta \in [-1,0]$ correspond to concept suppression, with $\delta=-1$ yielding complete suppression within the edited region, while $\delta>0$ amplifies the concept activation.
% The localized intervention is propagated through the learned bridge, yielding an updated global approximation
% $z'_{\mathrm{G},\mathrm{app}} =
% \mathrm{ReLU}\!\left(\mathcal{C}\bar{z}'_{\mathrm{L}} + b\right)$,
% where $\bar{z}'_{\mathrm{L}}=\phi(Z'_{\mathrm{L}})$ denotes the recomputed pooled local activation.
% The updated model prediction is then obtained by feeding $z'_{\mathrm{G},\mathrm{app}}$ to the classifier:
% $\ell'(I) =
% W_{\mathrm{cls}} z'_{\mathrm{G},\mathrm{app}} + b_{\mathrm{cls}}$.
The intervention is propagated through the learned bridge, yielding an updated global approximation
$z'_{\mathrm{G},\mathrm{app}} =
\mathrm{ReLU}\!\left(\mathcal{C}\bar{z}'_{\mathrm{L}} + b\right)$,
where $\bar{z}'_{\mathrm{L}}$ denotes the recomputed pooled local activation.
We estimate the additive change of the global concepts as $\Delta = z'_{\mathrm{G},\mathrm{app}} - z_{\mathrm{G},\mathrm{app}}$, and transfer these changes to the original global representation, i.e.,
$z'_{\mathrm{G}}= \mathrm{ReLU}(z_{\mathrm{G}}+\Delta)$. The updated prediction is then computed as
$\ell'(I)=W_{\mathrm{cls}}z'_{\mathrm{G}}+b_{\mathrm{cls}}$.
% where $\bar{z}'_{\mathrm{L}}$ denotes the recomputed pooled local activation.
% We estimate each global concept's relative change as $\alpha_i = \frac{z'_{\mathrm{G},\mathrm{app}}[i]}
%       {z_{\mathrm{G},\mathrm{app}}[i]}$,
% with $\alpha_i=1$ for zero $z_{\mathrm{G},\mathrm{app}}$ entries, and transfer these changes to the original global representation:
% $z'_{\mathrm{G}}=\alpha\odot z_{\mathrm{G}}$,
% where $\odot$ denotes element-wise multiplication. The updated prediction is then computed as
% $\ell'(I)=W_{\mathrm{cls}}z'_{\mathrm{G}}+b_{\mathrm{cls}}$.

% #################################################
%                   EXPERIMENTS
% #################################################

\section{Experiments and Results}
\label{experiments}

We evaluate the different components of our framework: compositional structure recovery, LG-CBM spatial concept grounding and classification, and multiscale model steering. We further validate interpretability and concept naming via a user study.
While our framework is model-agnostic, our experimental evaluation primarily focuses on ViT-based architectures.
Additional results are provided in the supplementary, including LG-SAE evaluation using standard SAE metrics, experiments with a DINOv2 backbone, and additional quantitative and qualitative results.

\subsection{Compositional Structure Recovery}
\label{sec:composition_eval}
We evaluate the ability of the bridge matrix $\mathcal{C}$ to \textit{learn} meaningful local--global compositional relations from real-world data compared to post-hoc statistical associations.
% A key contribution of LG-SAE is the ability of the bridge matrix $\mathcal{C}$ to \textit{learn} genuine compositional connections between local and global concepts during training, rather than merely reflecting statistical co-occurrence patterns computed post hoc.
% To quantitatively evaluate this capability, we design an experiment that measures how well the connections encoded in $\mathcal{C}$ align with a ground-truth ontology from a real-world dataset containing human annotations of Local--Global relationships.

\textbf{Dataset and Ground-Truth Ontology.}
We evaluate our method on the Part-ImageNet dataset~\citep{he2022partimagenet}, a subset of ImageNet that includes pixel-level annotations of object parts alongside image-level object category labels.
The dataset has a hierarchical structure: each image belongs to one of 11 super-categories (e.g., \textit{Bird}), each of which is associated with a fixed set of semantic object parts (e.g., \textit{Bird} $\to$ {\textit{Head, Body, Wing, Foot, Tail}}).
Let $\mathcal{G}$ denote the set of 11 super-categories, $\mathcal{P}$ the set of all annotated parts, and $\mathcal{I}_g$ the set of images belonging to super-category $g$. We define the ground-truth ontology matrix as
\begin{equation}
    \mathcal{C}_{\mathrm{gt}}(g, p) = \begin{cases} 1 & \text{if } p \text{ is an annotated semantic part of super-category } g, \\ 0 & \text{otherwise,} \end{cases}
    \quad g \in \mathcal{G},\ p \in \mathcal{P}.
\end{equation}
This defines a bipartite ground-truth graph over the node sets $\mathcal{G}$ and $\mathcal{P}$, with $|\mathcal{G}|\cdot|\mathcal{P}|$ possible edges and
$\sum_{g \in \mathcal{G}} |\mathcal{P}_g|$
positive edges, where $\mathcal{P}_g$ denotes the set of object parts associated with super-category $g$.

\textbf{Evaluation Protocol.}
We train LG-SAE on top of CLIP ViT-B/16~\citep{radford2021learning} using the ImageNet training set, as described in Section~\ref{training_process}. We use CLIP's final image embedding as the global representation and patch tokens from its penultimate layer as local representations. For the local and global streams of LG-SAE, we experiment with both TopK and BatchTopK SAE variants. We use a dictionary size of $d_{\text{L}}=8192$ with $K=12$ for the local SAE and $d_{\text{G}}=2048$ with $K=64$ for the global SAE.
See Appendix~\ref{sec:composition_implementation_details} for full implementation and hyperparameter selection details.

% We train LG-SAE on CLIP ViT-B/16~\cite{radford2021learning} using ImageNet, with the final image embedding as the global representation and penultimate-layer patch tokens as local representations. We experiment with both TopK and BatchTopK configurations, using a dictionary size of $d_{\mathrm{L}}=8192$ with $K=12$ for the local SAE and $d_{\mathrm{G}}=2048$ with $K=64$ for the global SAE. See Appendix~\ref{sec:composition_implementation_details} for full implementations details.

\emph{Step 1: Concept anchoring.}
Since LG-SAE is trained on the entire ImageNet dataset with large dictionary sizes, only a subset of its features correspond to the ground-truth ontology $\mathcal{C}_{\mathrm{gt}}$ of Part-ImageNet. Therefore, we first identify LG-SAE features that best correspond to Part-ImageNet's super-categories and parts.
For each super-category $g \in \mathcal{G}$ we find the global feature that is most selectively activated by that category. Specifically, we measure the point-biserial correlation~\citep{kornbrot2014point} $r_{\mathrm{pb}}(i, g)$ between activations of each global feature $z_{\text{G}}[i]$ and the binary indicator of the image category $g$, and choose the most selective feature, i.e.,
$i^*(g) = \operatorname*{arg\,max}_{i \in \{1,\ldots,d_{\text{G}}\}} r_{\mathrm{pb}}(i, g)$.
Similarly, for each part $p \in \mathcal{P}$ we identify the local feature whose spatial activations most align with the part's segmentation mask. We compute the per-image  point-biserial correlation $r_{\mathrm{pb}}(j,p)$ between the local feature activations $Z_{\text{L}}[:,j]$ and the binary segmentation mask of the part $p$, and average the scores over all images containing part $p$.
We then select the most aligned local feature,
$j^*(p) = \operatorname*{arg\,max}_{j \in \{1,\ldots,d_{\text{L}}\}}
    \frac{1}{|\mathcal{I}_p|}\sum_{\mathcal{I}_p} r_{\mathrm{pb}}(j,p)$,
% $j^*(p)
% =
% \operatorname*{arg\,max}_{j \in \{1,\ldots,d_{\mathrm{L}}\}}
% \frac{1}{|\mathcal{I}_p|}
% \sum_{n\in\mathcal{I}_p}
% r_{\mathrm{pb}}^{\,n}(j,p),$
where $\mathcal{I}_p$ is the set of all images containing part $p$.

\emph{Step 2: Subgraph extraction.}
Using the anchored features
$\{i^*(g)\}_{g \in \mathcal{G}}$ and
$\{j^*(p)\}_{p \in \mathcal{P}}$, we extract the corresponding entries from $\mathcal{C}$ and binarize them using a threshold $\tau$. This yields a subgraph that can be directly compared with $\mathcal{C}_{\mathrm{gt}}$, i.e.,
${\mathcal{C}^*}(g, p) = \mathbf{1}\bigl[\mathcal{C}_{i^*(g),\, j^*(p)} > \tau\bigr]$.
For LG-SAE we set $\tau=0$, since training enforces both non-negativity and sparsity in $\mathcal{C}$.

\textbf{Baselines.}
We compare our learnable bridge approach with two post-hoc statistical baselines that construct alternative bridge matrices from the same global and local activations of LG-SAE, and use them as drop-in replacements for $\mathcal{C}$.

(1)~\textit{Co-occurrence:}
% Following~\citet{wittenmayer2026cfm}, we score each global--local feature pair $(i,j)$ by the fraction of local feature $j$'s total activation that occurs in images where global feature $i$ is active:
Following~\cite{wittenmayer2026cfm}, we score each global--local feature pair $(i,j)$ by the fraction of local feature $j$'s activation occurring in images where global feature $i$ is active:
\begin{equation}
D_\mathrm{co}(i,j)
=
\frac{
\sum_{n=1}^{N} \mathbbm{1}[z^n_{\mathrm{G}}[i] > 0] \cdot \phi(Z^n_{\mathrm{L}})[j]
}{
\sum_{n=1}^{N} \bar{z}^{\,n}_{\mathrm{L}}[j] + \epsilon
},
\end{equation}
where $\epsilon$ is a small constant added for numerical stability.

(2)~\textit{PPMI:}
% As a second association-based baseline,
We score each feature pair $(i,j)$ using Positive Pointwise Mutual Information (PPMI):
\begin{equation}
D_\mathrm{ppmi}(i,j)
=
\max\left(0,\log \frac{P(i,j)}{P(i)P(j)}\right).
\end{equation}
The marginal probabilities $P(i)$ and $P(j)$ and the joint probability $P(i,j)$ are estimated from the global activations $z^n_{\mathrm{G}}[i]$ and pooled local activations $\phi(Z^n_{\mathrm{L}})[j]$ over the training set. For each baseline, we apply the same subgraph extraction procedure and select $\tau$ such that the resulting matrix matches the sparsity of the learned $\mathcal{C}$, ensuring a fair comparison.

\begin{table}[t]
\centering
\caption{\textbf{Compositional structure recovery on Part-ImageNet.}
Best results in each configuration are shown in bold.}
\label{tab:composition_recovery_clip}

\setlength{\tabcolsep}{3pt}
\renewcommand{\arraystretch}{1.1}
\small

\newcommand{\meanstd}[2]{%
    \ensuremath{#1\,{\scriptstyle\pm\,#2}}}

\begin{tabularx}{\linewidth}{@{}l*{6}{>{\centering\arraybackslash}X}@{}}
\toprule
& \multicolumn{3}{c}{\textbf{TopK}}
& \multicolumn{3}{c}{\textbf{BatchTopK}} \\
\cmidrule(lr){2-4}
\cmidrule(lr){5-7}

\textbf{Method}
& \textbf{Precision} & \textbf{Recall} & \textbf{F$_1$}
& \textbf{Precision} & \textbf{Recall} & \textbf{F$_1$} \\
\midrule

PPMI
& \textbf{0.909} & 0.250 & 0.392
& \textbf{0.923} & 0.300 & 0.453 \\

Co-occurrence
& 0.885 & 0.575 & 0.697
& 0.833 & 0.625 & 0.714 \\

\rowcolor{blue!15}
LG-SAE (ours)
& \meanstd{0.896}{0.005}
& \meanstd{\mathbf{0.647}}{0.01}
& \meanstd{\mathbf{0.751}}{0.005}
& \meanstd{0.897}{0.003}
& \meanstd{\mathbf{0.889}}{0.01}
& \meanstd{\mathbf{0.893}}{0.006} \\

\bottomrule
\end{tabularx}
\end{table}

% \begin{table}[t]
% \centering
% \caption{\textbf{Compositional structure recovery on Part-ImageNet.}
% Best results in each configuration are shown in bold.}
% \label{tab:composition_recovery_clip}

% \setlength{\tabcolsep}{3.5pt}
% \renewcommand{\arraystretch}{1.05}

% \begin{tabular}{lcccccc}
% \toprule
% & \multicolumn{3}{c}{\textbf{TopK}}
% & \multicolumn{3}{c}{\textbf{BatchTopK}} \\
% \cmidrule(lr){2-4}
% \cmidrule(lr){5-7}
% \textbf{Method}
% & \textbf{Precision} & \textbf{Recall} & \textbf{F$_1$}
% & \textbf{Precision} & \textbf{Recall} & \textbf{F$_1$} \\
% \midrule

% PPMI
% & \textbf{0.909} & 0.250 & 0.392
% & \textbf{0.923} & 0.300 & 0.453 \\

% Co-occurrence
% & 0.885 & {0.575} & {0.697}
% & 0.833 & {0.625} & {0.714} \\

% \rowcolor{blue!15}
% LG-SAE (ours)
% & ${0.896 \pm {\scriptstyle 0.005}}$ & $\mathbf{0.647 \pm {\scriptstyle 0.01}}$ & $\mathbf{0.755 \pm {\scriptstyle 0.005}}$
% & ${0.897 \pm {\scriptstyle 0.003}}$ & $\mathbf{0.889 \pm {\scriptstyle 0.01}}$ & $\mathbf{0.893 \pm {\scriptstyle 0.006}}$ \\

% \bottomrule
% \end{tabular}
% \end{table}

\begin{figure}[t]
    \centering
    \includegraphics[width=\linewidth]{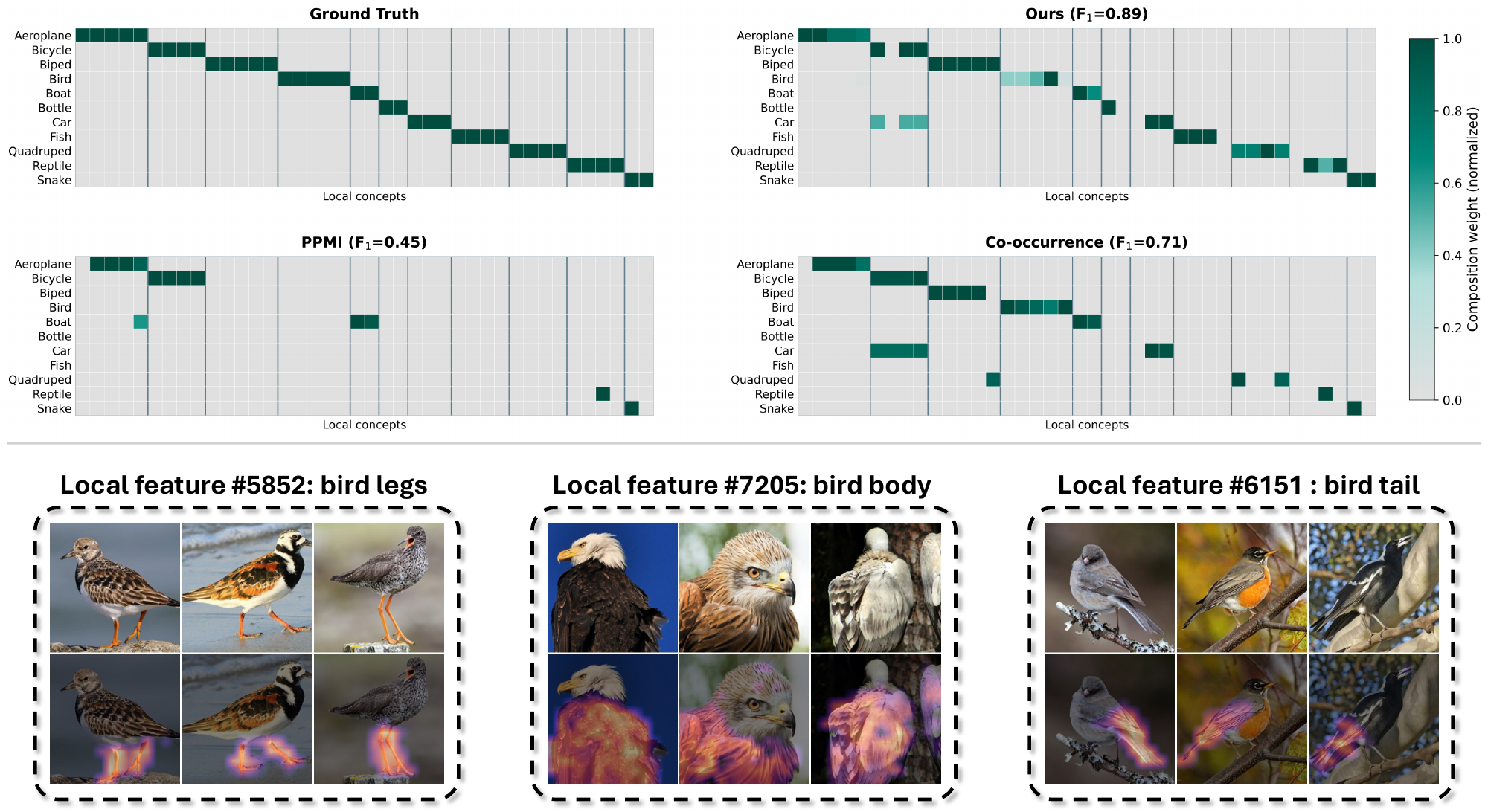}
    \caption{\textbf{Compositional graphs of LG-SAE vs. statistical methods.} Top: LG-SAE captures cleaner compositional relations compared with PPMI and co-occurrence baselines. Bottom: examples of LG-SAE local features aligned with semantic parts of the ``bird'' super-category.}
    \label{fig:C_matrix_compare}
\end{figure}

\textbf{Results.}
We compare each extracted subgraph with $\mathcal{C}_{\mathrm{gt}}$ using precision, recall, and F$_1$ score. Figure~\ref{fig:C_matrix_compare} visualizes the resulting graphs.
As shown in Table~\ref{tab:composition_recovery_clip}, LG-SAE achieves the highest F$_1$ score for both SAE configurations, primarily due to substantially better recall while maintaining high precision. The co-occurrence baseline achieves lower precision and recall than LG-SAE, while PPMI exhibits high precision but substantially lower recall, missing many valid part--whole links. These results demonstrate that learning the bridge matrix during training recovers the ground-truth compositional structure more effectively than post-hoc activation statistics.

\begin{wraptable}[8]{r}{0.42\textwidth}
% \vspace{-10pt}
\centering
\caption{Ablation study.}
\label{tab:C_ablation}
%\scriptsize
\setlength{\tabcolsep}{1.8pt}
\renewcommand{\arraystretch}{0.9}
\begin{tabular}{cccccc}
\toprule
$\mathcal{L}_{\mathrm{rec}}$ &
$\mathcal{L}_{\mathrm{align}}$ &
Iter. & Pre. & Rec. & F$_1$ \\
\midrule
\checkmark &  &  & 0.722 & 0.650 & 0.684 \\
 & \checkmark &  & 0.926 & 0.625 & 0.746 \\
\checkmark & \checkmark &  & \textbf{0.964} & 0.675 & 0.794 \\
\checkmark & \checkmark & \checkmark & 0.897 & \textbf{0.889} & \textbf{0.893} \\
\bottomrule
\end{tabular}
\vspace{-12pt}
\end{wraptable}

\textbf{Ablations.}
\label{sec:composition_ablation}
We ablate the components used to learn the compositional bridge in Table~\ref{tab:C_ablation}. Using only $\mathcal{L}_{\mathrm{rec}}$ yields moderate performance, showing that reconstructing global representations alone is insufficient for accurate concept alignment. Using only $\mathcal{L}_{\mathrm{align}}$ gives high precision but lower recall, suggesting that it captures selective yet incomplete associations. Combining both losses improves performance, reflecting their complementary roles: $\mathcal{L}_{\mathrm{align}}$ promotes feature-level consistency, while $\mathcal{L}_{\mathrm{rec}}$ preserves semantic fidelity in the model's representation space. Finally, applying the iterative refinement training stage further increases recall and overall F$_1$, yielding the best balance between precision and recall.

\subsection{LG-CBM Evaluation}
\label{sec:LG_CBM_eval}

We evaluate LG-SAE as a multiscale concept bottleneck layer by assessing: (i)~local feature grounding on zero-shot segmentation benchmarks; and (ii)~global feature quality through downstream classification accuracy. We use the BatchTopK LG-SAE trained on CLIP, as described in Section~\ref{sec:composition_eval}. 
For both evaluations, we report the mean and standard deviation across three independently trained models with different random seeds.
Full implementation details and results with DINOv2 backbone are provided in the supplementary.

\textbf{Spatial concept grounding.}
We evaluate LG-CBM local concept grounding quality, following the zero-shot concept grounding protocol of CFM~\citep{wittenmayer2026cfm}. Experiments are conducted on Part-ImageNet~\citep{he2022partimagenet} and COCO-Stuff~\citep{caesar2018coco}, two hierarchical datasets with image-level object labels and pixel-level annotations. Following~\citep{wittenmayer2026cfm}, we treat local feature activations $Z_{\mathrm L}$ as spatial grounding maps and compare them with ground-truth masks using three metrics: \textit{Locality}, which measures spatial alignment with semantic masks; \textit{Consistency}, which measures whether the same local features activate for the same semantic concepts across images; and \textit{Impurity}, which measures the extent to which activations remain concentrated within specific semantic regions. Higher values are better for Locality and Consistency, whereas lower values are better for Impurity.
Table~\ref{tab:segmentations_results} compares LG-CBM with recent concept localization methods, including SALF-CBM~\citep{benou2025show}, PatchSAE~\citep{lim2024sparse}, and CFM~\citep{wittenmayer2026cfm}. LG-CBM outperforms these baselines across all reported metrics, indicating that its local concepts are both semantically meaningful and spatially precise. 
% These results suggest that learning local features jointly with an explicit compositional structure yields more faithful and interpretable spatial grounding.

\begin{table*}[t]
\centering
\caption{\textbf{Concept grounding}. Best results are in bold. Baselines taken from~\cite{wittenmayer2026cfm}.}
\label{tab:segmentations_results}

\begin{adjustbox}{max width=\textwidth}
\begin{tabular}{lcccccc}
\toprule
& \multicolumn{2}{c}{\textbf{Locality} $\uparrow$}
& \multicolumn{2}{c}{\textbf{Consistency} $\uparrow$}
& \multicolumn{2}{c}{\textbf{Impurity} $\downarrow$} \\
\cmidrule(lr){2-3}
\cmidrule(lr){4-5}
\cmidrule(lr){6-7}
\textbf{Method} & IN-Part & COCO-Stuff & IN-Part & COCO-Stuff & IN-Part & COCO-Stuff \\
\midrule
SALF-CBM & 16.4 & 19.5 & 9.9 & 6.7 & 0.841 & 1.143 \\
% DN-CBM   & 18.7 & 17.9 & 9.3 & 5.7 & 0.761 & 1.235 \\
% MSAE     & 20.9 & 20.0 & 10.6 & 7.9 & 0.833 & 1.198 \\
PatchSAE & 32.2 & 34.6 & 13.1 & 12.7 & 0.353 & 0.663 \\
CFM      & 44.3 & 44.5 & 15.3 & 13.1 & 0.333 & 0.539 \\
\midrule
\rowcolor{blue!15}
\textsc{LG-CBM} (ours) &
$\mathbf{49.1 \pm {\scriptstyle 0.74}}$ &
$\mathbf{50.7 \pm {\scriptstyle 0.51}}$ &
$\mathbf{21.8 \pm {\scriptstyle 0.8}}$ &
$\mathbf{21.1 \pm {\scriptstyle 0.83}}$ &
$\mathbf{0.206 \pm {\scriptstyle 0.01}}$ &
$\mathbf{0.406 \pm {\scriptstyle 0.01}}$ \\
\bottomrule
\end{tabular}
\end{adjustbox}

\end{table*}

\begin{wraptable}{r}{0.5\columnwidth}
% \vspace{5pt}
\centering
\caption{\textbf{Classification accuracy}. Best results are in bold.
Baselines taken from~\cite{wittenmayer2026cfm}.}
\label{tab:clip_classification_results}
\setlength{\tabcolsep}{3pt}
\resizebox{\linewidth}{!}{%
\begin{tabular}{lccc}
\toprule
\textbf{Method} & \textbf{Grounding} & \textbf{ImageNet} & \textbf{Places} \\
\midrule
Linear Probe & -- & 80.2 & 55.1 \\
\midrule
LF-CBM~\citep{oikarinen2023label}
    & \textcolor{red}{\textbf{\ding{55}}} & 75.4 & 50.6 \\
LaBo~\citep{yang2023language}
    & \textcolor{red}{\textbf{\ding{55}}} & 78.9 & -- \\
CDM~\citep{panousis2023sparse}
    & \textcolor{red}{\textbf{\ding{55}}} & 79.3 & 52.6 \\
DCLIP~\citep{menon2022visual}
    & \textcolor{red}{\textbf{\ding{55}}} & 68.0 & 40.3 \\
DN-CBM~\citep{rao2024discover}
    & \textcolor{red}{\textbf{\ding{55}}} & \textbf{79.5} & 55.1 \\
D-CBM~\citep{prasse2024dcbm}
    & \textcolor{red}{\textbf{\ding{55}}} & 70.5 & 50.9 \\
SALF-CBM~\citep{benou2025show}
    & \textcolor{green}{$\checkmark$} & 76.3 & 49.4 \\
CFM~\citep{wittenmayer2026cfm}
    & \textcolor{green}{$\checkmark$} & 78.9 & \textbf{55.4} \\
\midrule
\rowcolor{blue!15}
\textsc{LG-CBM} (ours)
    & \textcolor{blue}{$\checkmark$}
    & $78.4 \pm {\scriptstyle 0.1}$
    & $54.5 \pm {\scriptstyle 0.2}$ \\
\bottomrule
\end{tabular}%
}
\end{wraptable}

% \begin{wraptable}{r}{0.5\columnwidth}
% % \vspace{-5pt}
% \centering
% \caption{\textbf{Classification accuracy}. Best results are in bold.
% Baselines taken from~\cite{wittenmayer2026cfm}.}
% \label{tab:clip_classification_results}
% \setlength{\tabcolsep}{3pt}
% \resizebox{\linewidth}{!}{%
% \begin{tabular}{lccc}
% \toprule
% \textbf{Method} & \textbf{Grounding} & \textbf{ImageNet} & \textbf{Places} \\
% \midrule
% Linear Probe & -- & 80.2 & 55.1 \\
% \midrule
% LF-CBM~\citep{oikarinen2023label}
%     & \textcolor{red}{\textbf{\ding{55}}} & 75.4 & 50.6 \\
% LaBo~\citep{yang2023language}
%     & \textcolor{red}{\textbf{\ding{55}}} & 78.9 & -- \\
% CDM~\citep{panousis2023sparse}
%     & \textcolor{red}{\textbf{\ding{55}}} & 79.3 & 52.6 \\
% DCLIP~\citep{menon2022visual}
%     & \textcolor{red}{\textbf{\ding{55}}} & 68.0 & 40.3 \\
% DN-CBM~\citep{rao2024discover}
%     & \textcolor{red}{\textbf{\ding{55}}} & \textbf{79.5} & 55.1 \\
% D-CBM~\citep{prasse2024dcbm}
%     & \textcolor{red}{\textbf{\ding{55}}} & 70.5 & 50.9 \\
% SALF-CBM~\citep{benou2025show}
%     & \textcolor{green}{\checkmark} & 76.3 & 49.4 \\
% CFM~\citep{wittenmayer2026cfm}
%     & \textcolor{green}{\checkmark} & 78.9 & \textbf{55.4} \\
% \midrule
% \rowcolor{blue!15}
% \textsc{LG-CBM} (ours)
%     & \textcolor{blue}{\checkmark} & 78.4 \pm {\scriptstyle 0.1} & 54.5 \pm {\scriptstyle 0.2} \\
% \bottomrule
% \end{tabular}%
% }
% \vspace{-6pt}
% \end{wraptable}

\textbf{Classification accuracy.}
% \vspace{2pt}
We evaluate classification performance on two large-scale datasets:
ImageNet and Places365. For each dataset, we train LG-SAE using
the same BatchTopK configuration as in Section~\ref{sec:composition_eval}.
We then train a linear classification layer on top of its global features, as described in Section~\ref{LG-CBMs}, and evaluate classification accuracy using the corresponding test set.
Table~\ref{tab:clip_classification_results} shows the results of LG-CBM compared to several existing CBM methods using the same backbone model. LG-CBM achieves competitive performance with the highest-performing CBMs, demonstrating its ability to provide high-quality
multiscale explanations while preserving downstream performance.

\subsection{Multiscale Model Steering}
\label{sec:model_steering}

We evaluate the multiscale steerability of LG-CBM through a controlled concept-suppression experiment. We use the Part-ImageNet validation set, which enables us to suppress local concepts within precise semantic regions defined by ground-truth part masks and measure the resulting effect on the model's final prediction. We use the same LG-CBM model evaluated in Section~\ref{sec:LG_CBM_eval}.

\textbf{Evaluation protocol.}
We evaluate steering in three steps: concept anchoring, computation of pre-steering activations, and localized concept suppression.

(1)~\textit{Concept anchoring}: Following Section~\ref{sec:composition_eval}, for each super-category $g$ and part $p$ in the dataset, we identify the best-aligned global feature $i^*(g)$ and local feature $j^*(p)$, respectively.

(2)~\textit{Pre-steering activations}: For each image, we compute the local concept activations $Z_{\mathrm{L}}[p,j]$, pooled activations $\bar{z}_{\mathrm{L}}=\phi(Z_{\mathrm{L}})$, approximated global concept activation $z_{\mathrm{G},\mathrm{app}} = \mathrm{ReLU}\!\left(\mathcal{C}\bar{z}_{\mathrm{L}} + b\right)$, and classification logits $\ell = W_{\mathrm{cls}} z_{\mathrm{G}} + b_{\mathrm{cls}}$,
as described in Section~\ref{steerability}.

(3)~\textit{Local concept suppression}: We suppress the anchored local concept $j^*(p)$ within the ground-truth mask $M$ corresponding to part $p$, using suppression strengths $\delta \in \{0, -0.2, -0.4, \ldots, -1.0\}$. Here, $\delta = 0$ corresponds to no intervention, while $\delta = -1$ corresponds to complete suppression within the mask. We propagate each intervention through the model and measure the resulting drop in the predicted-class probability for each (image, part) pair:
$\Delta p(\delta) = p(c^*) - p'(c^*)$,
where $p(c) = \operatorname{softmax}(\ell)_c$, $p'(c)=\operatorname{softmax}(\ell')_c$, and $c^* = \arg\max_c p(c)$.

\textbf{Baselines.}
Under identical suppression conditions, we compare our results to two baselines:

(1)~\textit{Outside-mask steering}: To evaluate spatial specificity, we apply the same suppression of $j^*(p)$ to the spatial complement of the ground-truth mask, i.e., $1-M$.

(2)~\textit{Random-concept steering}: To evaluate concept specificity, we suppress $K=10$ 
randomly selected non-anchored local concepts and average the resulting probability drops. The concepts are sampled once per (image, part) pair and kept fixed across all $\delta$ values.

\begin{figure}[t]
    \centering
    \begin{subfigure}{0.43\linewidth}
        \centering
        \caption{Steering results on Part-ImageNet.}
        \label{fig:steering_suppression_results}
        \includegraphics[width=\linewidth]{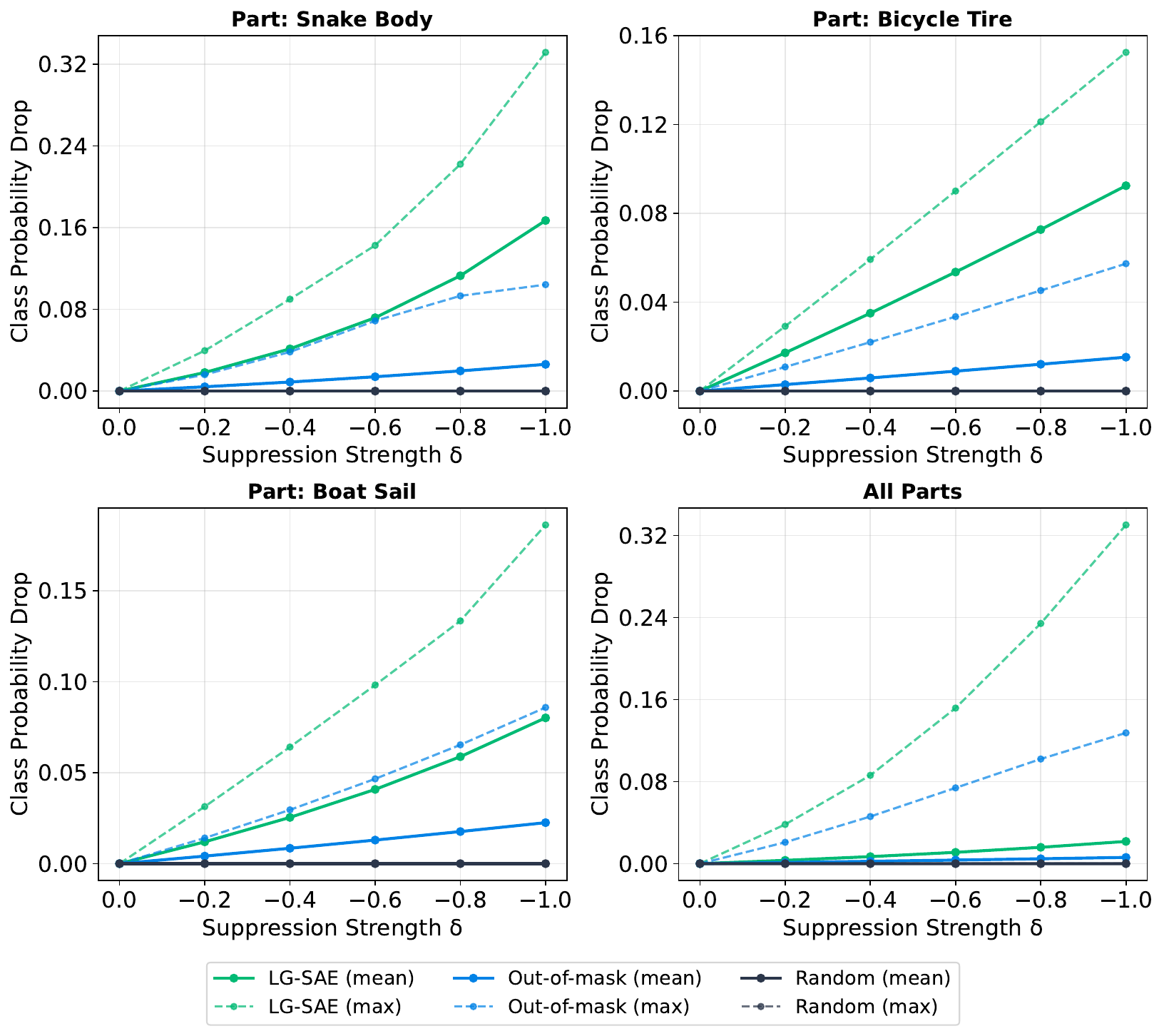}
    \end{subfigure}
    \hfill
    \begin{subfigure}{0.51\linewidth}
        \centering
        \caption{Misclassification correction example.}
        \label{fig:steering_example}
        \includegraphics[width=\linewidth]{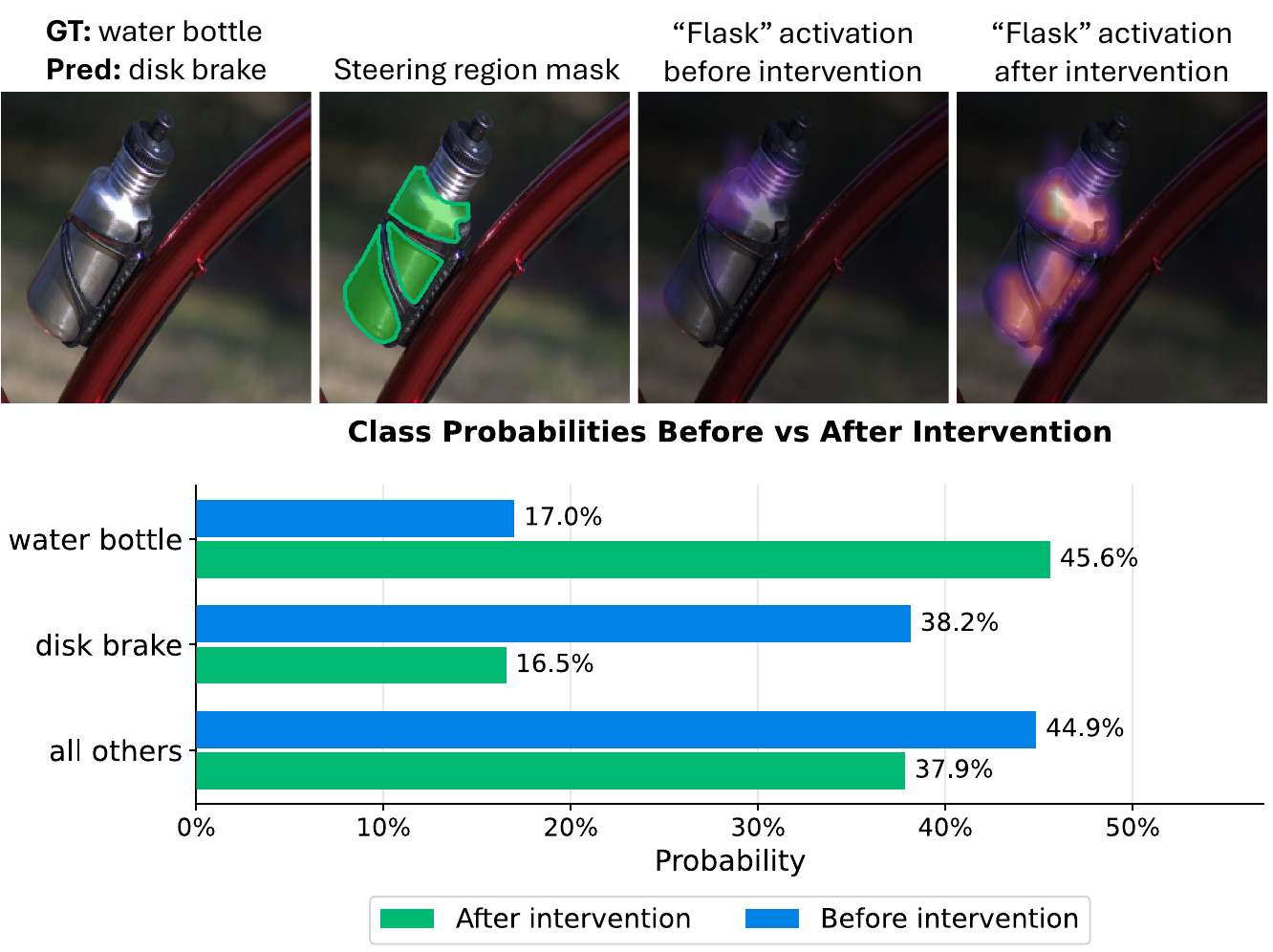}
    \end{subfigure}
    \caption{\textbf{Multiscale model steering results}. Left: local concept suppression on Part-ImageNet. Right: example of a local intervention resulting in a corrected model prediction.}
    \label{fig:steering_results}
\end{figure}

\textbf{Results.}
Figure~\ref{fig:steering_suppression_results} shows the mean and maximum drop in predicted-class probability across all images containing each of the three most steerable semantic parts (i.e., those whose suppression has the largest effect on the predicted-class probability), along with the mean and maximum drop across all parts in the dataset.
The results show that suppressing local concepts within semantically relevant regions leads to a monotonic drop in predicted-class probability as suppression strength increases.
Notably, suppressing a single anchored local concept out of 8,192 candidates, within its corresponding spatial region, has a substantial effect on the model's prediction. For example, the ``snake body'' concept exhibits a mean drop of up to 16.7\% and a maximum drop of over 30\%.
Moreover, suppressing concepts within their ground-truth masks has a substantially larger effect than suppressing them outside these regions, demonstrating the spatial specificity of our method.
In contrast, suppressing randomly selected concepts has virtually no effect on the model's predictions across all parts, further validating the concept specificity of our method.

\textbf{Misclassification correction example.}
In addition to concept suppression, LG-CBM enables localized concept amplification. Figure~\ref{fig:steering_example} shows an example in which the model initially misclassifies an image of a bicycle \textit{water bottle} as a \textit{disk brake}. By amplifying a single local concept corresponding to \textit{``Flask''} within the bottle region, the model's prediction shifts toward the correct ground-truth class.
This example illustrates how localized concept interventions provide an intuitive mechanism for steering LG-CBM predictions through semantically meaningful concept edits.

\subsection{User Study}
\label{sec:user_study}

We conduct a user study involving 32 participants to evaluate the human interpretability of LG-SAE's compositional graph and spatially aware concept naming. The full details of the study and example questions are provided in the supplementary.
% We recruited 32 participants for two anonymous online surveys evaluating concept composition and naming. Participation was voluntary and uncompensated, with no personal or sensitive data collected.

\begin{figure}[t]
\centering
\begin{subfigure}{0.48\linewidth}
\centering
\caption{Concept composition survey results.}
\label{fig:composition_survey_results}
\includegraphics[width=\linewidth]{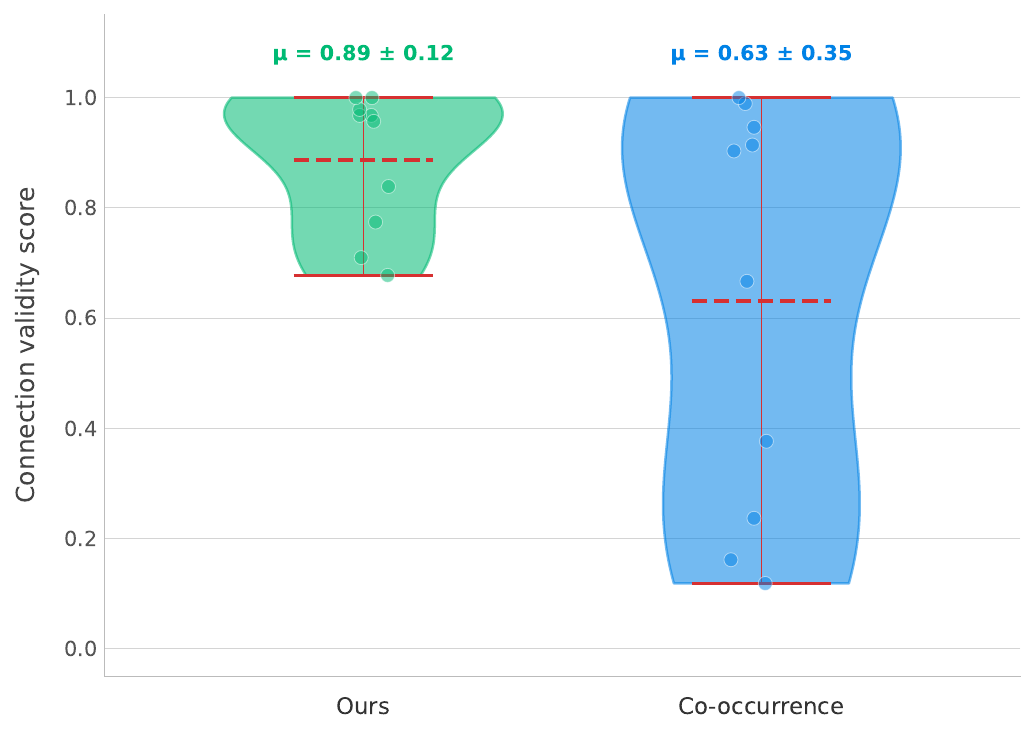}
\end{subfigure}
\hfill
\begin{subfigure}{0.48\linewidth}
\centering
\caption{Concept naming survey results.}
\label{fig:naming_survey_results}
\includegraphics[width=\linewidth]{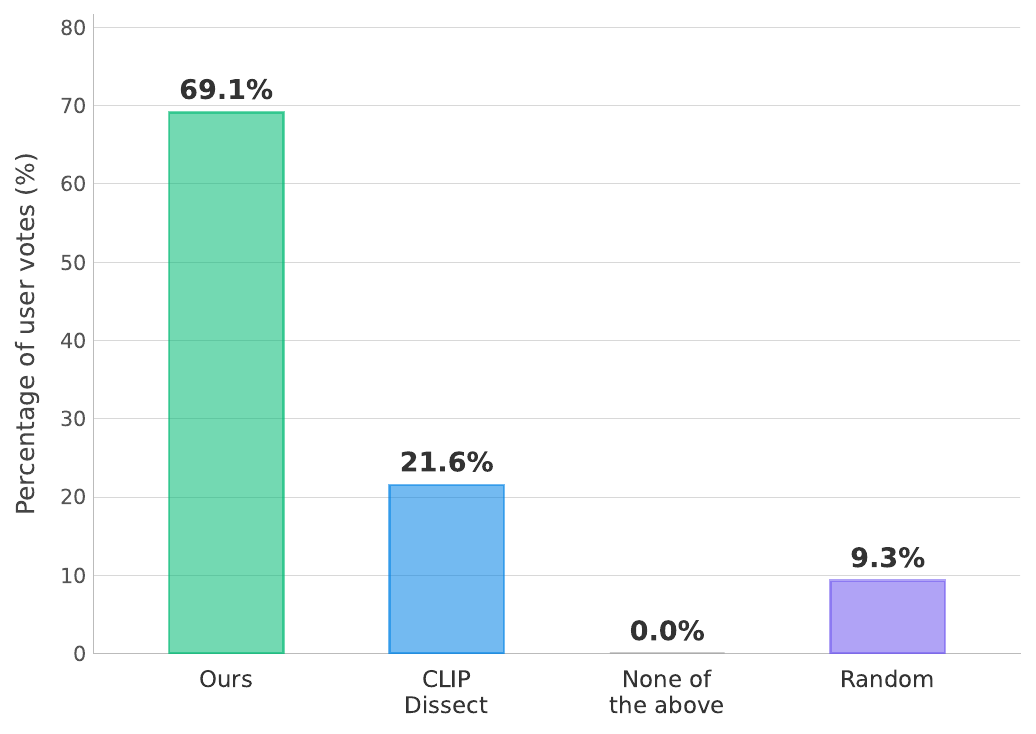}
\end{subfigure}
\caption{\textbf{User study results}.}
\label{fig:user_study_results}
\end{figure}

\paragraph{Concept composition survey.}
To evaluate the learned compositional bridge $\mathcal{C}$, we compare LG-SAE with the statistical co-occurrence baseline from Section~\ref{sec:composition_eval}. We rank global LG-SAE features by their point-biserial selectivity to ImageNet class labels and randomly sample two sets of ten concepts from the top 25\%, one per method. For each global feature $i$, we randomly select a connected local feature $j$ from the corresponding matrix: $\mathcal{C}$ for LG-SAE or $D_{\mathrm{co}}$ for the baseline.
For each pair $(i,j)$, participants view the three most activating images and corresponding spatial heatmaps. Given the global concept label, they provide a binary ``Yes/No'' judgment on whether the highlighted region represents a valid part of that concept. Responses are averaged across participants to obtain a connection validity score in $[0,1]$ per pair.
% Instructions and an example question appear in Figs.~\ref{fig:survey_1_instructions} and~\ref{fig:survey_1_question}.
Figure~\ref{fig:composition_survey_results} shows that LG-SAE yields more reliable compositional links than co-occurrence, with higher average validity and lower variance.

\paragraph{Concept naming survey.}
We compare our spatially aware naming method, presented in Section~\ref{concept_naming}, with CLIP-Dissect~\cite{oikarinen2022clip}. We use both methods to label LG-SAE's local features using the Broden vocabulary. We randomly select 15 features assigned different labels by the two methods. For each feature, participants view the five most activating images and corresponding feature activation maps, then select the label best describing the highlighted regions from four options: our label, the CLIP-Dissect label, a random label, or ``None of the above''. 
% Instructions and an example appear in Figs.~\ref{fig:survey_2_instructions} and~\ref{fig:survey_2_question}.
As shown in Figure~\ref{fig:naming_survey_results}, our labels receive higher ratings from participants compared to CLIP-Dissect, suggesting that incorporating spatial information better captures localized part-level semantics and reduces context-driven labeling errors.

% #################################################
%                   Conclusions
% #################################################
\section{Conclusion}
\label{conclusion}
We revisit an often overlooked question in visual interpretability: \emph{how are localized features organized into image-level semantic concepts?}
To address this question, we introduced LG-SAE, a multiscale interpretability framework that makes this cross-scale organization explicit by learning local and global concepts together with a sparse bridge that connects them.
% Our experiments show that LG-SAE captures meaningful compositional structure and supports spatially grounded explanations, spatially aware concept naming, and localized model steering.
Across our evaluations, the learned bridge recovers known part--whole structure more faithfully than post-hoc association methods. Building on this representation, LG-CBM provides spatially grounded concept explanations while maintaining competitive classification performance, and our spatially aware naming method improves concept interpretability. The representation also enables targeted intervention: editing a concept within a selected region produces specific, interpretable changes in the model's output. Together, these capabilities connect multiscale concept discovery, explanation, and model steering.

A limitation of the current framework is that the reliability of the learned bridge ultimately depends on the monosemanticity of the underlying SAE features, as entangled features may yield less reliable connections.
Future work may address this limitation through improved feature disentanglement and adaptive bridge-pruning. More broadly, we hope this work encourages a shift from interpreting visual features in isolation toward understanding the structures they form across scales.

% #################################################
%                  ACKNOWLEDGMENTS
% #################################################
\newpage

\begin{ack}
This work was supported in part by the Israel Ministry of Innovation, Science and Technology (MOST, 0007311).
\end{ack}

% #################################################
%                    REFERENCES
% #################################################

% \newpage

% \section*{References}
\bibliographystyle{unsrt}
\bibliography{refs}

%%%%%%%%%%%%%%%%%%%%%%%%%%%%%%%%%%%%%%%%%%%%%%%%%%%%%%%%%%%%

% \newpage
% \section*{References}

% References follow the acknowledgments in the camera-ready paper. Use unnumbered first-level heading for
% the references. Any choice of citation style is acceptable as long as you are
% consistent. It is permissible to reduce the font size to \verb+small+ (9 point)
% when listing the references.
% Note that the Reference section does not count towards the page limit.
% \medskip

% {
% \small

% [1] Alexander, J.A.\ \& Mozer, M.C.\ (1995) Template-based algorithms for
% connectionist rule extraction. In G.\ Tesauro, D.S.\ Touretzky and T.K.\ Leen
% (eds.), {\it Advances in Neural Information Processing Systems 7},
% pp.\ 609--616. Cambridge, MA: MIT Press.

% [2] Bower, J.M.\ \& Beeman, D.\ (1995) {\it The Book of GENESIS: Exploring
%   Realistic Neural Models with the GEneral NEural SImulation System.}  New York:
% TELOS/Springer--Verlag.

% [3] Hasselmo, M.E., Schnell, E.\ \& Barkai, E.\ (1995) Dynamics of learning and
% recall at excitatory recurrent synapses and cholinergic modulation in rat
% hippocampal region CA3. {\it Journal of Neuroscience} {\bf 15}(7):5249-5262.
% }

%%%%%%%%%%%%%%%%%%%%%%%%%%%%%%%%%%%%%%%%%%%%%%%%%%%%%%%%%%%%
\newpage
\appendix

\section*{Supplementary material}

\section{LG-SAE training}
\begin{algorithm}
\caption{Three-Stage Training of LG-SAE}
\label{alg:lg_sae_training}
\begin{algorithmic}[1]

%% ── Inputs / Outputs ─────────────────────────────────────────────────────────
\Require Global data $x_{\text{G}}$, local data $x_{\text{L}}$,
         reconstruction weight $\alpha$,
         sparsity regularizer $\lambda_{\mathcal C}$,
         refinement weight $\beta$
\Ensure  Trained global SAE $(E_{\text{G}}, D_{\text{G}})$,
         local SAE $(E_{\text{L}}, D_{\text{L}})$,
         bridge mapping $(\mathcal{C}, b)$

\Statex
%% ── Stage 1 ──────────────────────────────────────────────────────────────────
\Statex \textbf{Stage 1: Independent SAE Training} \Comment{Multiscale concept discovery}
\State Train $(E_{\text{G}}, D_{\text{G}})$ by minimizing $\mathcal{L}_{\mathrm{sae},\text{G}}(x_{\text{G}})$
\State Train $(E_{\text{L}}, D_{\text{L}})$ by minimizing $\mathcal{L}_{\mathrm{sae},\text{L}}(x_{\text{L}})$
\State Freeze parameters of $(E_{\text{G}}, D_{\text{G}})$ and $(E_{\text{L}}, D_{\text{L}})$

\Statex
%% ── Stage 2 ──────────────────────────────────────────────────────────────────
\Statex \textbf{Stage 2: Bridge Learning} \Comment{Learning compositional relations}
\State Encode inputs: $z_{\text{G}} \leftarrow E_{\text{G}}(x_{\text{G}})$,\quad $z_{\text{L}} \leftarrow E_{\text{L}}(x_{\text{L}})$
\State Pool local features: $\bar{{z}}_{\text{L}} \leftarrow \phi(Z_{\text{L}})$
\State Approximate global features from locals: 
% $\hat{z}_{\text{G}} \leftarrow f_{\mathcal{C},b}(z_{\text{L}})$
$z_{\mathrm{G},\mathrm{app}} \leftarrow \text{ReLU}(\mathcal{C} \bar{{z}}_{\text{L}} + b)$
\State Optimize $(\mathcal{C}, b)$ by minimizing:
\[
  \mathcal{L}_{\mathrm{bridge}}
  \;=\;
  \underbrace{\|z_{\text{G}} - z_{\mathrm{G},\mathrm{app}}\|_2^2}_{\text{feature alignment}}
  \;+\;
  \underbrace{\alpha\,\|x_{\text{G}} - D_{\text{G}}(z_{\mathrm{G},\mathrm{app}})\|_2^2}_{\text{reconstruction}}
  \;+\;
  \underbrace{\lambda_{\mathcal C}\,\mathcal{R}(\mathcal{C})}_{\text{regularization}}
\]

\Statex
%% ── Stage 3 ──────────────────────────────────────────────────────────────────
\Statex \textbf{Stage 3: Alternating Refinement} \Comment{Structure-aware finetuning}
\Repeat
   \State \textit{(a) Refine local SAE:} freeze $(\mathcal{C}, b)$; update $(E_{\text{L}}, D_{\text{L}})$ by minimizing:
    \[
      \mathcal{L}_{\text{refine}} = \mathcal{L}_{\text{sae,L}} + \beta \|z_\text{G} - z_{\mathrm{G},\mathrm{app}}\|_2^2
    \]
    % \State Recompute $z_G \leftarrow E_G(x_G)$ with updated encoder
    \State \textit{(b) Refine bridge:} freeze $(E_{\text{L}}, D_{\text{L}})$; update $(\mathcal{C}, b)$ by minimizing $\mathcal{L}_{\mathrm{bridge}}$
    % \State Recompute $\hat{z}_{\text{G}} \leftarrow f_{\mathcal{C},b}(z_{\text{L}})$ with updated bridge
\Until{convergence}
\Statex
\State \Return $(E_{\text{G}}, D_{\text{G}})$,\; $(E_{\text{L}}, D_{\text{L}})$,\; $(\mathcal{C}, b)$

\end{algorithmic}
\end{algorithm}

\clearpage
%%%%%%%%%%%%%%%%%%%%%%%%%%%%%%%%%%%%%%%%%%%%%%%%%%%%%%%%%%%%

\section{Implementation details}
\label{sec:implementation_details}
\paragraph{Computational resources.}
All experiments in the paper were performed using a single NVIDIA RTX A6000 GPU with 48~GB of memory.

\subsection{Local--Global Compositional Structure}
\label{sec:composition_implementation_details}
The implementation details for the experiment described in Section~\ref{sec:composition_eval} are as follows.

\textbf{LG-SAE training.}
We train LG-SAE on top of the CLIP ViT-B/16 vision encoder using the ImageNet training set, as described in Section~\ref{training_process}. We use patch tokens from its penultimate layer as local representations and CLIP's image embeddings (a projection of the final \texttt{CLS} token) as global representations. We experiment with two widely used SAE configurations for both the local and global streams: TopK and BatchTopK.

We use different dictionary sizes for the local and global SAEs since patch tokens and image-level representations differ in semantic granularity. The local stream captures a heterogeneous set of spatially localized patterns, motivating a larger dictionary, whereas the global stream captures aggregated image-level semantics, motivating a smaller one. We set the local dictionary size to $d_{\text{L}}=8192$ to align with CFM~\citep{wittenmayer2026cfm}, and use a global dictionary one-quarter the size, $d_{\text{G}}=2048$.
We select the sparsity levels by sweeping $K$ for each SAE and examining the reconstruction--sparsity trade-off as shown in Figure~\ref{fig:rec_vs_sparsity}. We choose values near the elbow regions, beyond which larger values of $K$ yield marginal reconstruction improvements at the cost of reducing sparsity, resulting in $K_{\text{L}}=12$ and $K_{\text{G}}=64$.
These values preserve most of the reconstruction fidelity while maintaining sparse, interpretable features.

For all configurations, training was done using Adam optimizer. In stage 1, we train the local SAE for 2 epochs on individual patch tokens with a learning rate of $5e^{-4}$ and auxiliary loss weight of $\alpha_{aux}=0.01$. We train the global SAE on CLIP image embeddings for 120 epochs with the same learning rate and auxiliary loss weight. 
In stage 2, we use average pooling for $\phi$ and train the bridge parameters for 5 epochs with a learning rate of $1e^{-3}$, reconstruction loss weight of $\alpha=0.5$ and $L_1$ regularization strength of $\lambda_{\mathcal{C}}=3$.
In stage 3 (iterative refinement), we use $\beta=0.5$ and lower the local SAE learning rate to $1e^{-4}$, while other training parameters are kept as in previous stages.

\textbf{PPMI computation.}
Given a dataset with \(N\) images, let ${z}_{\mathrm{G},n}[i]$ denote the activation of global concept $i$ on image $n$, and  $\bar{z}_{\mathrm{L},n}[j]$ denote the pooled activation of local concept $j$ on that image.
We define per-image binary activation indicators for each of them as:
\[
a_{n,i}^{g}
=
\mathbf{1}\!\left[{z}_{\mathrm{G},n}[i] > 0\right],
\qquad
a_{n,j}^{\ell}
=
\mathbf{1}\!\left[\bar{z}_{\mathrm{L},n}[j] > 0\right].
\]

The marginal activation probabilities are estimated as the fraction of images
in which each concept is active:
\[
P(i)
=
\frac{1}{N}\sum_{n=1}^{N} a_{n,i}^{g},
\qquad
P(j)
=
\frac{1}{N}\sum_{n=1}^{N} a_{n,j}^{\ell}.
\]

The joint activation probability is estimated as the fraction of images in
which both concepts are simultaneously active:
\[
P(i,j)
=
\frac{1}{N}\sum_{n=1}^{N}
a_{n,i}^{g} a_{n,j}^{\ell}.
\]

The Pointwise Mutual Information (PMI) is then:
\[
\operatorname{PMI}(i,j)
=
\log\left(
\frac{P(i,j)}
     {P(i)P(j)}
\right).
\]

Finally, the Positive Pointwise Mutual Information (PPMI) bridge matrix is
defined as:
\[
\operatorname{PPMI}(i,j)
=
\max\left(0,\operatorname{PMI}(i,j)\right).
\]

\subsection{LG-CBM Evaluations}
\label{sec:LG_CBM_implementation_details}
The implementation details for the experiment described in Section~\ref{sec:LG_CBM_eval} are as follows. For the classification accuracy experiment, we evaluate on two large-scale classification datasets: Places365 and ImageNet. For each dataset, we train LG-SAE on top of CLIP ViT-B/16 vision encoder, using a local BatchTopK SAE with dictionary size $d_L = 8192$ and $K = 12$, and a global BatchTopK SAE with $d_G = 2048$ and $K = 64$. We then train a linear classification layer on top of the LG-SAE global features and evaluate its accuracy on the corresponding test set. The classifier was trained for 100 epochs using a standard cross-entropy loss with a learning rate of $2e-4$ and weight decay regularization strength of $1e-4$.

For the spatial concept grounding experiment we use the LG-SAE trained on ImageNet with the same parameters as in the classification accuracy experiment. For both experiments, we report the mean and standard deviation across three independently trained models with different random seeds.

\clearpage
%%%%%%%%%%%%%%%%%%%%%%%%%%%%%%%%%%%%%%%%%%%%%%%%%%%%%%%%%%%%

\section{LG-SAE Evaluation}
\label{sec:LG-SAE_evaluation}
\paragraph{SAE metrics.}
We evaluate both the local and global SAEs using standard metrics: reconstruction fidelity ($R^2$), sparsity ($L_0$), and dead-feature ratio, for the TopK and BatchTopK LG-SAE configurations trained on CLIP backbone. Similarly, we evaluate the learned bridge using reconstruction fidelity of $z_{\mathrm{G}}$ from its approximation $z_{\mathrm{G},\mathrm{app}}$, and sparsity, measured as the fraction of zero-valued entries. As baselines, we report standard TopK and BatchTopK SAEs trained independently with the same hyperparameters.

Table~\ref{tab:sae_metrics_clip} shows that both LG-SAE configurations achieve high reconstruction fidelity and low dead-feature ratios across the local and global SAEs. Their performance closely matches that of the corresponding baselines, indicating that learning cross-scale feature relationships does not compromise the quality of the individual SAE representations.

% SAE METRICS - CLIP backbone
\begin{table}[h]
\centering
\caption{\textbf{SAE evaluation with CLIP backbone.}}
\label{tab:sae_metrics_clip}
\begin{tabular}{lcccccccc}
\toprule
\textbf{Model}
& \multicolumn{3}{c}{\textbf{Local SAE ($k=12$)}}
& \multicolumn{3}{c}{\textbf{Global SAE ($k=64$)}}
& \multicolumn{2}{c}{\textbf{Bridge}} \\
\cmidrule(lr){2-4}
\cmidrule(lr){5-7}
\cmidrule(lr){8-9}
& $\mathbf{R^2}$ & $\mathbf{L_0}$ & \textbf{Dead}
& $\mathbf{R^2}$ & $\mathbf{L_0}$ & \textbf{Dead}
& $\mathbf{R^2}$ & $\mathbf{Sparsity}$ \\
\midrule
TopK
& 0.911 & 12 & 0.1\%
& 0.947 & 64 & 0.0\%
& - & - \\

\cellcolor{blue!15}LG-SAE (TopK)
& \cellcolor{blue!15}0.911 & \cellcolor{blue!15}12 & \cellcolor{blue!15}0.14\%
& \cellcolor{blue!15}0.947 & \cellcolor{blue!15}64 & \cellcolor{blue!15}0.0\%
& \cellcolor{blue!15}0.809 & \cellcolor{blue!15}99.5\% \\

BatchTopK
& 0.913 & 12 & 0.3\%
& 0.948 & 64 & 0.0\%
& - & - \\

\cellcolor{blue!15}LG-SAE (BatchTopK)
& \cellcolor{blue!15}0.914 & \cellcolor{blue!15}12 & \cellcolor{blue!15}0.4\%
& \cellcolor{blue!15}0.948 & \cellcolor{blue!15}64 & \cellcolor{blue!15}0.0\%
& \cellcolor{blue!15}0.805 & \cellcolor{blue!15}99.5\% \\
\bottomrule
\end{tabular}
\end{table}

\begin{figure}[h]
    \centering
    \includegraphics[width=0.9\linewidth]{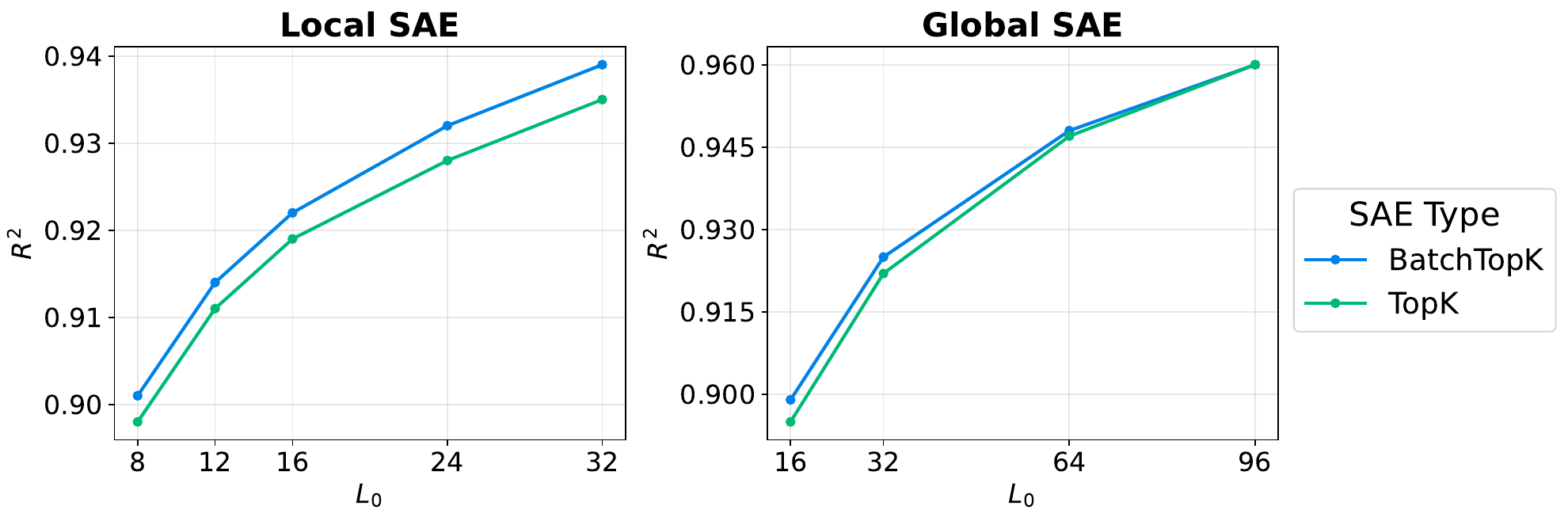}
    \caption{Reconstruction fidelity ($R^2$) as a function of sparsity ($L_0$) for the local and global SAEs.}
    \label{fig:rec_vs_sparsity}
\end{figure}

\paragraph{Reconstruction vs. sparsity trade-off.}
To guide the selection of $K$ across our experiments, we evaluate the reconstruction fidelity of the local and global LG-SAE streams over a range of sparsity levels for both the TopK and BatchTopK configurations.
Figure~\ref{fig:rec_vs_sparsity} shows diminishing returns in reconstruction fidelity ($R^2$) as $L_0$ increases, with an ``elbow'' around $L_0=12$--$16$ for the local SAE and $L_0=32$--$64$ for the global SAE.

\clearpage
%%%%%%%%%%%%%%%%%%%%%%%%%%%%%%%%%%%%%%%%%%%%%%%%%%%%%%%%%%%%
\section{Experiments with DINOv2 Backbone}
\label{sec:dinov2_results}

In addition to CLIP, we further evaluate our framework on the DINOv2 ViT-B/14 pretrained backbone~\cite{oquab2023dinov2}.

\paragraph{Implementation details.}
We train LG-SAE on top of DINOv2 ViT-B/14 following the same training procedure described in Section~\ref{training_process}.
To demonstrate that our method also works with same-layer representations, we use \texttt{CLS} and patch tokens from the penultimate layer (both with dimensionality $m=768$) as the global and local representations, respectively.
As in the CLIP experiments, we evaluate both TopK and BatchTopK configurations for the local and global SAEs. We preserve the dictionary expansion factors and activation-sparsity ratios used in our CLIP experiments, yielding $(d_G,K_G)=(3072,96)$ and $(d_L,K_L)=(8192,12)$.
Unless stated otherwise, training was performed on ImageNet using the same training parameters as described in~\ref{sec:composition_implementation_details}.

\subsection{LG-SAE Evaluation}
This experiment complements the LG-SAE evaluation performed on CLIP backbone in Appendix~\ref{sec:LG-SAE_evaluation}.
We evaluate the local and global SAEs, as well as the learned bridge, for both the TopK and BatchTopK LG-SAE configurations using the same metrics as in Appendix~\ref{sec:LG-SAE_evaluation}.
As baselines, we report standard TopK and BatchTopK SAEs trained independently with the same hyperparameters.

Table~\ref{tab:sae_metrics_dino} shows that both LG-SAE configurations achieve high reconstruction fidelity and low dead-feature ratios across the local and global SAEs. Their performance closely matches that of the corresponding baselines, indicating that learning cross-scale feature relationships does not compromise the quality of the individual SAE representations.

% DINOv2 RESULTS
\begin{table}[h]
\centering
\caption{\textbf{SAE evaluation with DINOv2 backbone.}}
\label{tab:sae_metrics_dino}
\begin{tabular}{lcccccccc}
\toprule
\textbf{Model}
& \multicolumn{3}{c}{\textbf{Local SAE ($k=12$)}}
& \multicolumn{3}{c}{\textbf{Global SAE ($k=96$)}}
& \multicolumn{2}{c}{\textbf{Bridge}} \\
\cmidrule(lr){2-4}
\cmidrule(lr){5-7}
\cmidrule(lr){8-9}
& $\mathbf{R^2}$ & $\mathbf{L_0}$ & \textbf{Dead}
& $\mathbf{R^2}$ & $\mathbf{L_0}$ & \textbf{Dead}
& $\mathbf{R^2}$ & $\mathbf{Sparsity}$ \\
\midrule
TopK
& 0.942 & 12 & 1.6\%
& 0.968 & 96 & 0.2\%
& - & - \\

\cellcolor{blue!15}LG-SAE (TopK)
& \cellcolor{blue!15}0.943 & \cellcolor{blue!15}12 & \cellcolor{blue!15}1.5\%
& \cellcolor{blue!15}0.968 & \cellcolor{blue!15}96 & \cellcolor{blue!15}0.2\%
& \cellcolor{blue!15}0.795 & \cellcolor{blue!15}98.6\% \\

BatchTopK
& 0.943 & 12 & 0.11\%
& 0.969 & 96 & 0.16\%
& - & - \\

\cellcolor{blue!15}LG-SAE (BatchTopK)
& \cellcolor{blue!15}0.943 & \cellcolor{blue!15}12 & \cellcolor{blue!15}0.19\%
& \cellcolor{blue!15}0.969 & \cellcolor{blue!15}96 & \cellcolor{blue!15}0.16\%
& \cellcolor{blue!15}0.798 & \cellcolor{blue!15}98.7\% \\
\bottomrule
\end{tabular}
\end{table}

\subsection{Compositional Structure Recovery}
This experiment complements the compositional structure recovery results on CLIP backbone from Section~\ref{sec:composition_eval}.
We follow the same evaluation protocol as in Section~\ref{sec:composition_eval}, and compare our results to the same co-occurrence and PPMI baselines.

Results are shown in Table~\ref{tab:composition_recovery_dino}. As in the CLIP experiment, LG-SAE achieves the highest F$_1$ score for both SAE configurations, with substantially better recall while maintaining high precision. These results demonstrate our framework's ability to successfully recover meaningful compositional structure across different backbone models.

\begin{table}[t]
\centering
\caption{\textbf{Compositional structure recovery on Part-ImageNet with DINOv2 backbone.}
Best results in each configuration are shown in bold.}
\label{tab:composition_recovery_dino}

\setlength{\tabcolsep}{3pt}
\renewcommand{\arraystretch}{1.1}
\small

\newcommand{\meanstd}[2]{%
    \ensuremath{#1\,{\scriptstyle\pm\,#2}}}

\begin{tabularx}{\linewidth}{@{}l*{6}{>{\centering\arraybackslash}X}@{}}
\toprule
& \multicolumn{3}{c}{\textbf{TopK}}
& \multicolumn{3}{c}{\textbf{BatchTopK}} \\
\cmidrule(lr){2-4}
\cmidrule(lr){5-7}

\textbf{Method}
& \textbf{Precision} & \textbf{Recall} & \textbf{F$_1$}
& \textbf{Precision} & \textbf{Recall} & \textbf{F$_1$} \\
\midrule

PPMI
& \textbf{1.0} & 0.262 & 0.415
& \textbf{1.0} & 0.28 & 0.437 \\

Co-occurrence
& {0.937} & {0.375} & {0.536}
& {0.882} & {0.375} & {0.526} \\

\rowcolor{blue!15}
LG-SAE (ours)
& \meanstd{0.7}{0.009}
& \meanstd{\mathbf{0.7}}{0.013}
& \meanstd{\mathbf{0.7}}{0.01}
& \meanstd{0.732}{0.008}
& \meanstd{\mathbf{0.75}}{0.011}
& \meanstd{\mathbf{0.741}}{0.009} \\

\bottomrule
\end{tabularx}
\end{table}

\subsection{LG-CBM Evaluation}
This experiment complements the LG-CBM evaluation with CLIP backbone from Section~\ref{sec:LG_CBM_eval}.

\paragraph{Spatial concept grounding.}
\label{grounding_results_dino}
As in Section~\ref{sec:LG_CBM_eval}, we use the BatchTopK configuration of LG-SAE and evaluate the local concept grounding quality of LG-CBM following the zero-shot concept grounding protocol in~\cite{wittenmayer2026cfm}. 

Compared with CLIP, Table~\ref{tab:segmentations_results_dino} shows that LG-CBM with a DINOv2 backbone achieves better consistency results on both datasets, and comparable locality and impurity. Overall, the two backbones provide similar grounding quality.

\begin{table*}[h]
\centering
\caption{\textbf{Concept grounding results.} Comparing CLIP and DINOv2 backbones.}
\label{tab:segmentations_results_dino}

\begin{adjustbox}{max width=\textwidth}
\begin{tabular}{lcccccc}
\toprule
& \multicolumn{2}{c}{\textbf{Locality} $\uparrow$}
& \multicolumn{2}{c}{\textbf{Consistency} $\uparrow$}
& \multicolumn{2}{c}{\textbf{Impurity} $\downarrow$} \\
\cmidrule(lr){2-3}
\cmidrule(lr){4-5}
\cmidrule(lr){6-7}
\textbf{Method} & IN-Part & COCO-Stuff & IN-Part & COCO-Stuff & IN-Part & COCO-Stuff \\
\midrule
% $\mathbf{21.8 \pm {\scriptstyle 0.8}}$ &
% $\mathbf{20.1 \pm {\scriptstyle 0.83}}$ &
\textsc{LG-CBM} (CLIP) &
${49.1 \pm {\scriptstyle 0.74}}$ &
$\mathbf{50.7 \pm {\scriptstyle 0.51}}$ &
${21.8 \pm {\scriptstyle 0.8}}$ &
${21.1 \pm {\scriptstyle 0.83}}$ &
${0.206 \pm {\scriptstyle 0.01}}$ &
${0.406 \pm {\scriptstyle 0.01}}$ \\
\textsc{LG-CBM} (DINOv2) &
$\mathbf{49.3 \pm {\scriptstyle 0.61}}$ &
${48.4 \pm {\scriptstyle 0.55}}$ &
$\mathbf{26.6 \pm {\scriptstyle 0.69}}$ &
$\mathbf{22.8 \pm {\scriptstyle 0.8}}$ &
$\mathbf{0.2 \pm {\scriptstyle 0.01}}$ &
$\mathbf{0.404 \pm {\scriptstyle 0.02}}$ \\
\bottomrule
\end{tabular}
\end{adjustbox}

\end{table*}

% \begin{table}[h]
% \centering
% \caption{\textbf{Concept grounding results with CLIP and DINOv2 backbones}.}
% \label{tab:segmentations_results_dino}
% \setlength{\tabcolsep}{4.5pt}
% \begin{tabular}{lcccccc}
% \toprule
% & \multicolumn{2}{c}{\textbf{Locality} $\uparrow$} & \multicolumn{2}{c}{\textbf{Consistency} $\uparrow$} & \multicolumn{2}{c}{\textbf{Impurity} $\downarrow$} \\
% \cmidrule(lr){2-3}
% \cmidrule(lr){4-5}
% \cmidrule(lr){6-7}
% \textbf{Method} & IN-Part & COCO-Stuff & IN-Part & COCO-Stuff & IN-Part & COCO-Stuff \\
% \midrule
% $\mathbf{49.1 \pm {\scriptstyle 0.74}}$ &
% $\mathbf{50.7 \pm {\scriptstyle 0.51}}$ &
% $\mathbf{32.6 \pm {\scriptstyle 0.8}}$ &
% $\mathbf{33.6 \pm {\scriptstyle 0.93}}$ &
% $\mathbf{0.206 \pm {\scriptstyle 0.01}}$ &
% $\mathbf{0.406 \pm {\scriptstyle 0.01}}$
% \textsc{LG-CBM (CLIP)} & $\mathbf{49.1 \pm {\scriptstyle 0.74}}$ & $\mathbf{50.7 \pm {\scriptstyle 0.51}}$ & $\mathbf{34.5 \pm {\scriptstyle 0.8}}$ & $\mathbf{31.1 \pm {\scriptstyle 0.93}}$ & $\mathbf{0.2 \pm {\scriptstyle 0.01}}$ & $\mathbf{0.404 \pm {\scriptstyle 0.01}}$ \\
% \textsc{LG-CBM (DINOv2)} & 46.5 & 46.65 & \textbf{51.6} & 43.9 & \textbf{0.089} & \textbf{0.192}
% & $\mathbf{49.3 \pm {\scriptstyle 0.74}}$ & $\mathbf{48.4 \pm {\scriptstyle 0.51}}$ & $\mathbf{34.5 \pm {\scriptstyle 0.8}}$ & $\mathbf{31.1 \pm {\scriptstyle 0.93}}$ & $\mathbf{0.2 \pm {\scriptstyle 0.01}}$ & $\mathbf{0.404 \pm {\scriptstyle 0.01}}$\\
% \bottomrule
% \end{tabular}
% \end{table}

\paragraph{Classification accuracy.}
\label{classification_results_dino}

Following the CLIP experiment in Section~\ref{sec:LG_CBM_eval}, we train LG-CBM on ImageNet and Places365 datasets, and evaluate its classification accuracy on the corresponding test set. As in Section~\ref{sec:LG_CBM_eval}, we use the BatchTopK configuration of LG-SAE, and train the linear classifier as described in Appendix~\ref{sec:LG_CBM_implementation_details}.

Table~\ref{tab:classification_results_dino} shows that LG-CBM with a DINOv2 backbone achieves classification accuracy comparable to that of CLIP, with only a slight decrease in performance on both datasets.
% 78.4 \pm {\scriptstyle 0.7} & 54.4 \pm {\scriptstyle 0.2}
\begin{table}[h!]
\centering
\caption{\textbf{Classification accuracy results.} Comparing CLIP and DINOv2 backbones.}
\label{tab:classification_results_dino}
\setlength{\tabcolsep}{4pt}
\begin{tabular}{lccc}
\toprule
\textbf{Method} & \textbf{Grounding} & \textbf{ImageNet} & \textbf{Places365} \\
\midrule
\textsc{LG-CBM (CLIP)}
    & \textcolor{blue}{$\checkmark$}
    & $\mathbf{78.4} \pm {\scriptstyle 0.1}$
    & $\mathbf{54.5} \pm {\scriptstyle 0.2}$ \\
\textsc{LG-CBM (DINOv2)}
    & \textcolor{blue}{$\checkmark$}
    & $77.8 \pm {\scriptstyle 0.2}$
    & $53.7 \pm {\scriptstyle 0.3}$ \\
\bottomrule
\end{tabular}
\end{table}

% \begin{table}[h!]
% \centering
% \caption{\textbf{Classification accuracy results.} Comparing CLIP and DINOv2 backbones.}
% \label{tab:classification_results_dino}
% \setlength{\tabcolsep}{4pt}
% \begin{tabular}{lccc}
% \toprule
% \textbf{Method} & \textbf{Grounding} & \textbf{ImageNet} & \textbf{Places365} \\
% \midrule
% \textsc{LG-CBM (CLIP)}
%     & \textcolor{blue}{\checkmark}
%     & \textbf{78.4} \pm {\scriptstyle 0.1} & \textbf{54.5} \pm {\scriptstyle 0.2} \\
% \textsc{LG-CBM (DINOv2)}
%     & \textcolor{blue}{\checkmark}
%     & 77.8 \pm {\scriptstyle 0.2} & 53.7 \pm {\scriptstyle 0.3} \\
% \bottomrule
% \end{tabular}
% \end{table}

\clearpage
%%%%%%%%%%%%%%%%%%%%%%%%%%%%%%%%%%%%%%%%%%%%%%%%%%%%%%%%%%%%

\section{User Study}
\label{sec:user_study_appendix}

We conduct a user study to quantitatively evaluate the interpretability of LG-SAE's compositional graph and spatially aware concept naming. We recruited 32 participants to complete two online surveys, one evaluating concept composition and the other evaluating concept naming. The study was conducted anonymously, with no collection of personal or sensitive data. Participation was voluntary, and no compensation was provided.

\begin{figure}[H]
    \centering
    \includegraphics[width=\linewidth]{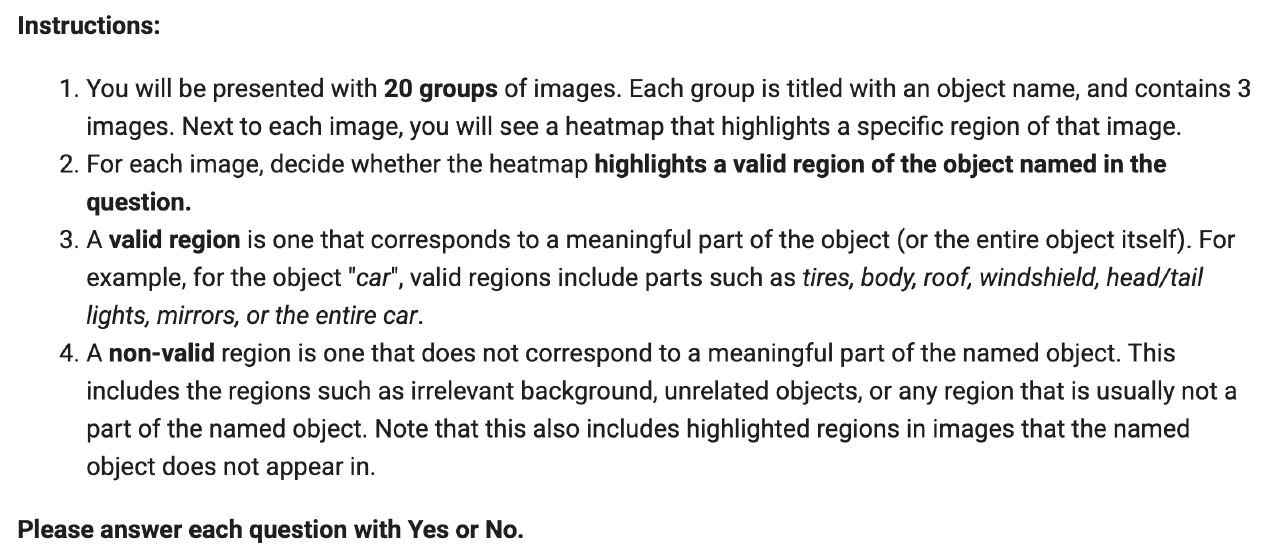}
    \caption{User instructions for concept composition survey.}
    \label{fig:survey_1_instructions}
\end{figure}

\begin{figure}[H]
    \centering
    \includegraphics[width=0.75\linewidth]{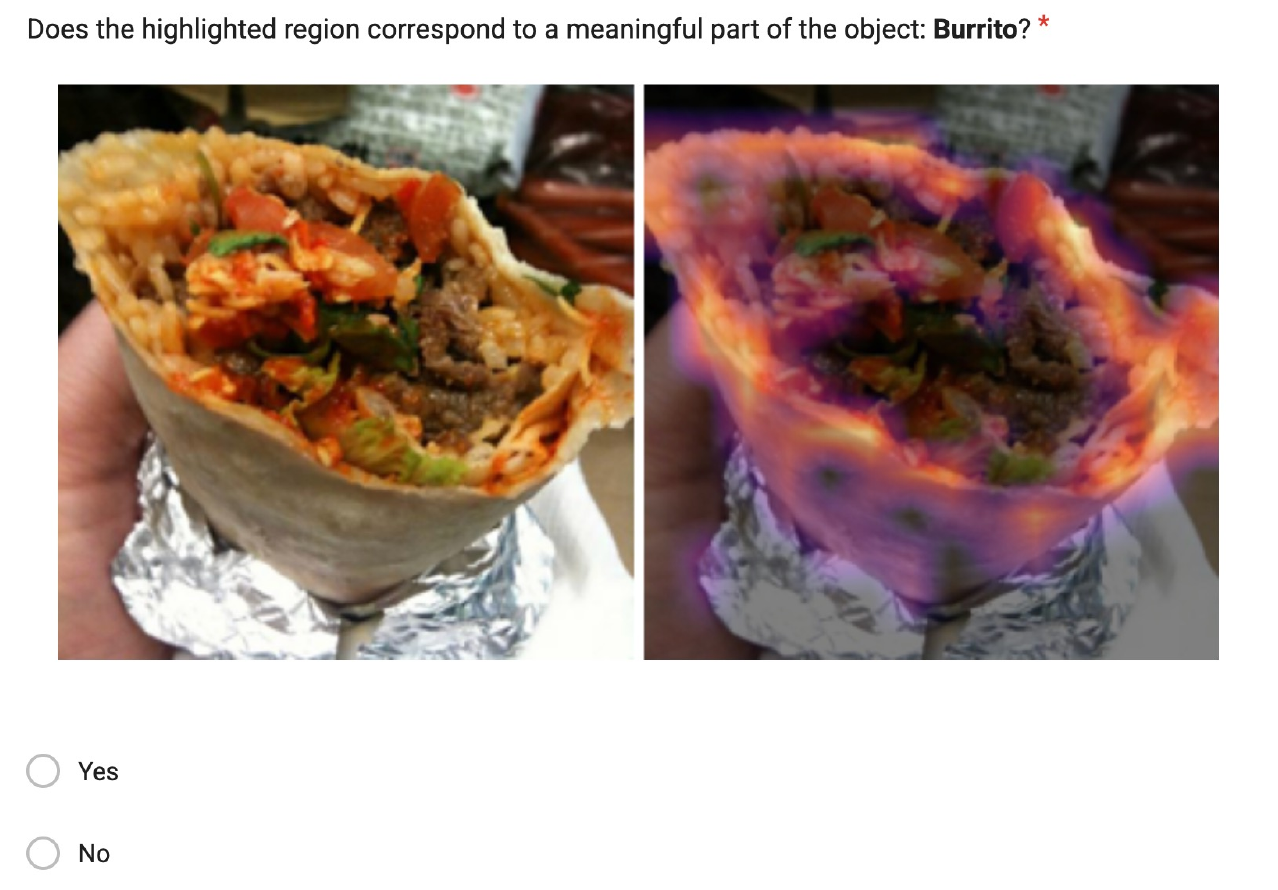}
    \caption{Example question for concept composition survey.}
    \label{fig:survey_1_question}
\end{figure}

\paragraph{Concept composition survey.}
Participants were instructed as shown in Fig.~\ref{fig:survey_1_instructions}. An example question is shown in Fig.~\ref{fig:survey_1_question}.

\clearpage

\begin{figure}[H]
    \centering
    \includegraphics[width=\linewidth]{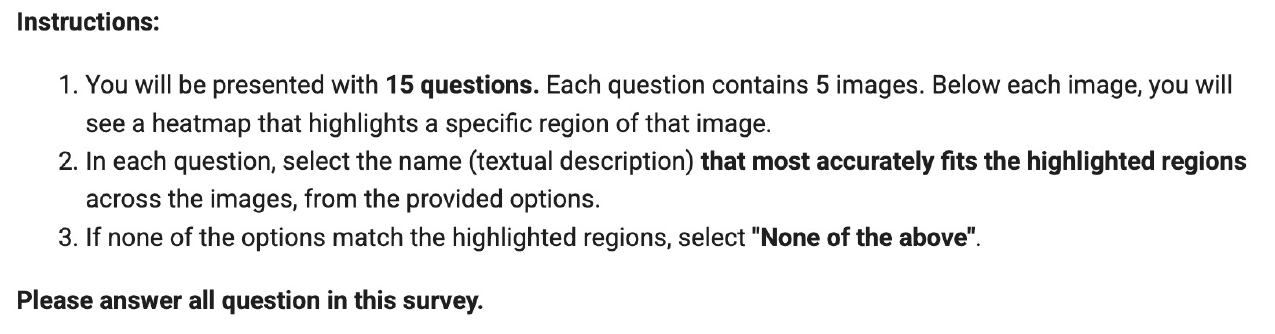}
    \caption{User instructions for concept naming survey.}
    \label{fig:survey_2_instructions}
\end{figure}

\begin{figure}[H]
    \centering
    \includegraphics[width=0.75\linewidth]{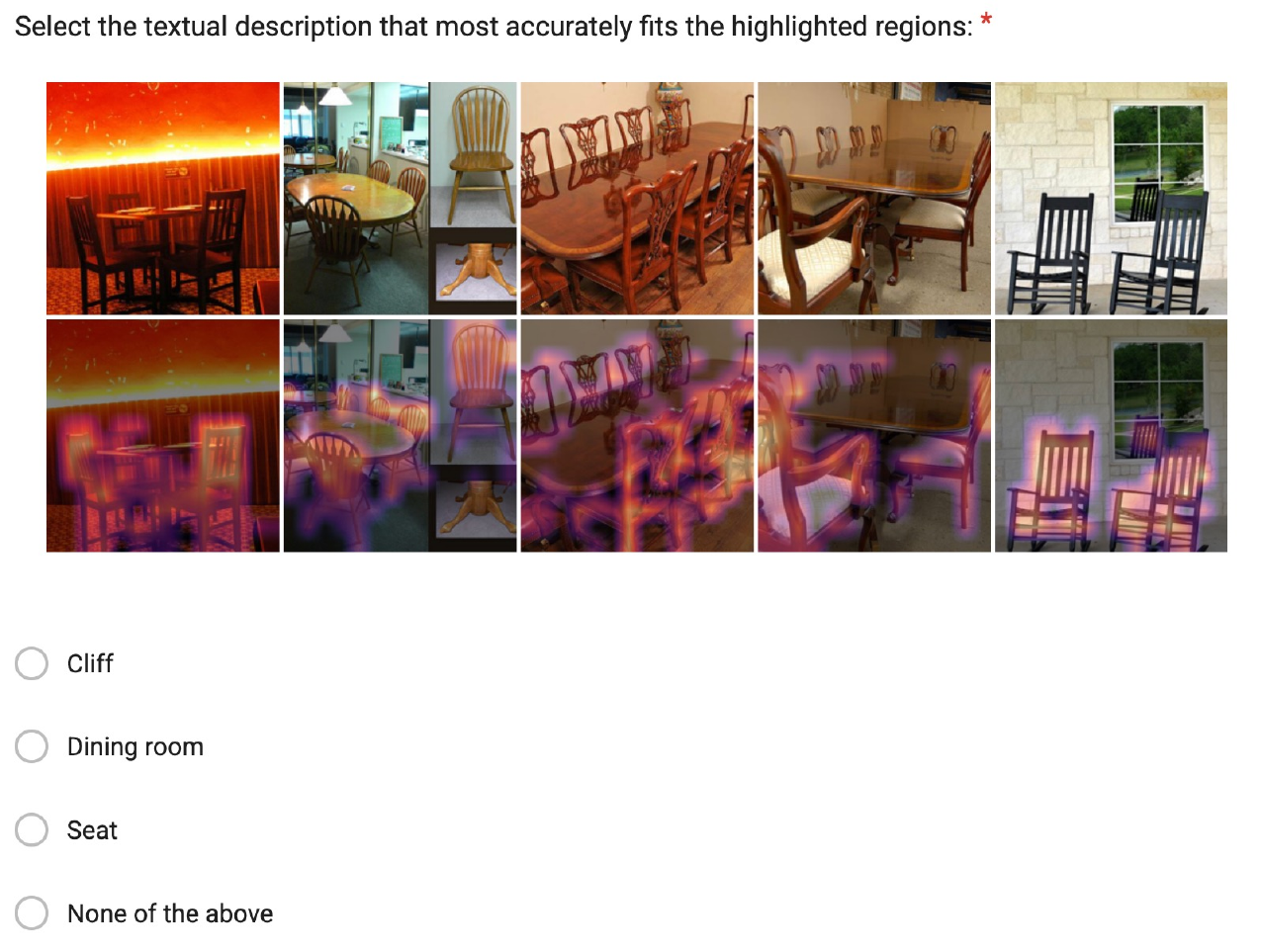}
    \caption{Example question for concept naming survey.}
    \label{fig:survey_2_question}
\end{figure}

\paragraph{Concept naming survey.}
Participants were instructed as shown in Fig.~\ref{fig:survey_2_instructions}. An example question is shown in Fig.~\ref{fig:survey_2_question}.

\clearpage

\section{Additional Quantitative Results}
\label{sec:quantitative_results}

\subsection{Concept Naming Evaluation}
In addition to the user study in Section~\ref{sec:user_study}, we provide a quantitative evaluation of our spatial concept naming method using Part-ImageNet labels as an oracle. Since Part-ImageNet provides both segmentation masks and semantic labels for object parts, the ground-truth mask of each of the 40 parts in the dataset can be treated as the spatial activation of a perfectly localized, monosemantic ``neuron'' whose true concept label is known (i.e., the name of the corresponding part, such as ``car side mirror''). We then evaluate how well the proposed spatial concept naming method recovers these known labels compared with the non-spatial CLIP-Dissect baseline.

For both methods, we use the same images, Soft-WPMI similarity function, and vocabulary, which contains the names of 11 super-categories and 40 parts (i.e., 51 possible concept labels). Each part (``oracle neuron'') is labeled using both methods.
We compare our method with CLIP-Dissect using two metrics: (1) \textbf{Accuracy}: the fraction of predicted labels that match the ground-truth label (higher is better); and (2) \textbf{Parent Confusion}: the fraction of predicted labels that match the part's parent super-category rather than the part itself (e.g., labeling a neuron corresponding to a car's side mirror simply as ``car''). This metric quantifies the failure mode in which a fine-grained part concept is collapsed into its coarse object category (lower is better).

Table~\ref{tab:concept_naming_oracle} shows that our method improves both accuracy and parent confusion compared to CLIP-Dissect. These results indicate that our method is able to better identify fine-grained concepts while substantially reducing contextual bias toward global object categories. Because this experiment uses oracle masks rather than learned SAE activations, it isolates the naming procedure from feature quality and provides objective evidence that the proposed spatial naming method more accurately identifies the concept represented by a localized feature.

\begin{table}[h]
    \centering
    \caption{Concept naming evaluation results.}
    \label{tab:concept_naming_oracle}
    \begin{tabular}{lcc}
        \toprule
        \textbf{Method} & \textbf{Accuracy} $\uparrow$ & \textbf{Parent confusion} $\downarrow$ \\
        \midrule
        CLIP-Dissect & 0.375 & 0.275 \\
        Ours & \textbf{0.45} & \textbf{0.2} \\
        \bottomrule
    \end{tabular}
\end{table}

% \begin{table}[h]
%     \centering
%     \caption{Concept naming evaluation results.}
%     \label{tab:concept_naming_oracle}
%     \begin{tabular}{lcc}
%         \toprule
%         \textbf{Method} & \textbf{Accuracy} \uparrow & \textbf{Parent confusion} \downarrow \\
%         \midrule
%         CLIP-Dissect  & 0.375 & 0.275 \\
%         Ours  & \textbf{0.45} & \textbf{0.2} \\
%         \bottomrule
%     \end{tabular}
% \end{table}

\subsection{Effect of spatial pooling operator}
\label{sec:pooling_results}
As described in Section~\ref{lg_sae}, the compositional bridge learns relations between global features $z_{\mathrm{G}}$ and pooled local activations
$\bar{z}_{\mathrm{L}}=\phi(Z_{\mathrm{L}})$, where $\phi$ is a spatial pooling operator.
We evaluate how different pooling operations affect compositional structure recovery. Specifically, we train BatchTopK LG-SAE following Appendix~\ref{sec:composition_implementation_details} using three pooling strategies: mean, max, and softmax pooling.
For softmax pooling, each entry is assigned the weight
$w_{p,j}=\exp(Z_{\mathrm{L}}[p,j])/\sum_{p'}\exp(Z_{\mathrm{L}}[p',j])$, and the pooled activation is defined as $\phi(Z_{\mathrm{L}})[j]=\sum_p w_{p,j}Z_{\mathrm{L}}[p,j]$.
We then repeat the Part-ImageNet compositional structure recovery experiment from Section~\ref{sec:composition_eval} using the same evaluation protocol. As shown in Table~\ref{tab:pooling_ablation}, mean-pooling achieves the best overall F$_1$ score. Although max- and softmax-pooling yield higher precision, they miss more valid connections, leading to substantially lower recall.

\begin{table}[h]
    \centering
    \caption{Effect of different pooling types on compositional structure recovery.}
    \label{tab:pooling_ablation}
    \begin{tabular}{lccc}
        \toprule
        \textbf{Pooling} & \textbf{Precision} & \textbf{Recall} & \textbf{F$_1$} \\
        \midrule
        Mean  & 0.897 & \textbf{0.889} & \textbf{0.893} \\
        Max  & 0.9 & 0.225 & 0.36 \\
        Softmax  & \textbf{1.0} & 0.525 & 0.689 \\
        \bottomrule
    \end{tabular}
\end{table}

\clearpage
%%%%%%%%%%%%%%%%%%%%%%%%%%%%%%%%%%%%%%%%%%%%%%%%%%%%%%%%%%%%

\section{Additional qualitative results}
\label{sec:qualitative_results}
\subsection{Spatial concept grounding}
\begin{figure}[H]
\centering
\includegraphics[width=\linewidth]{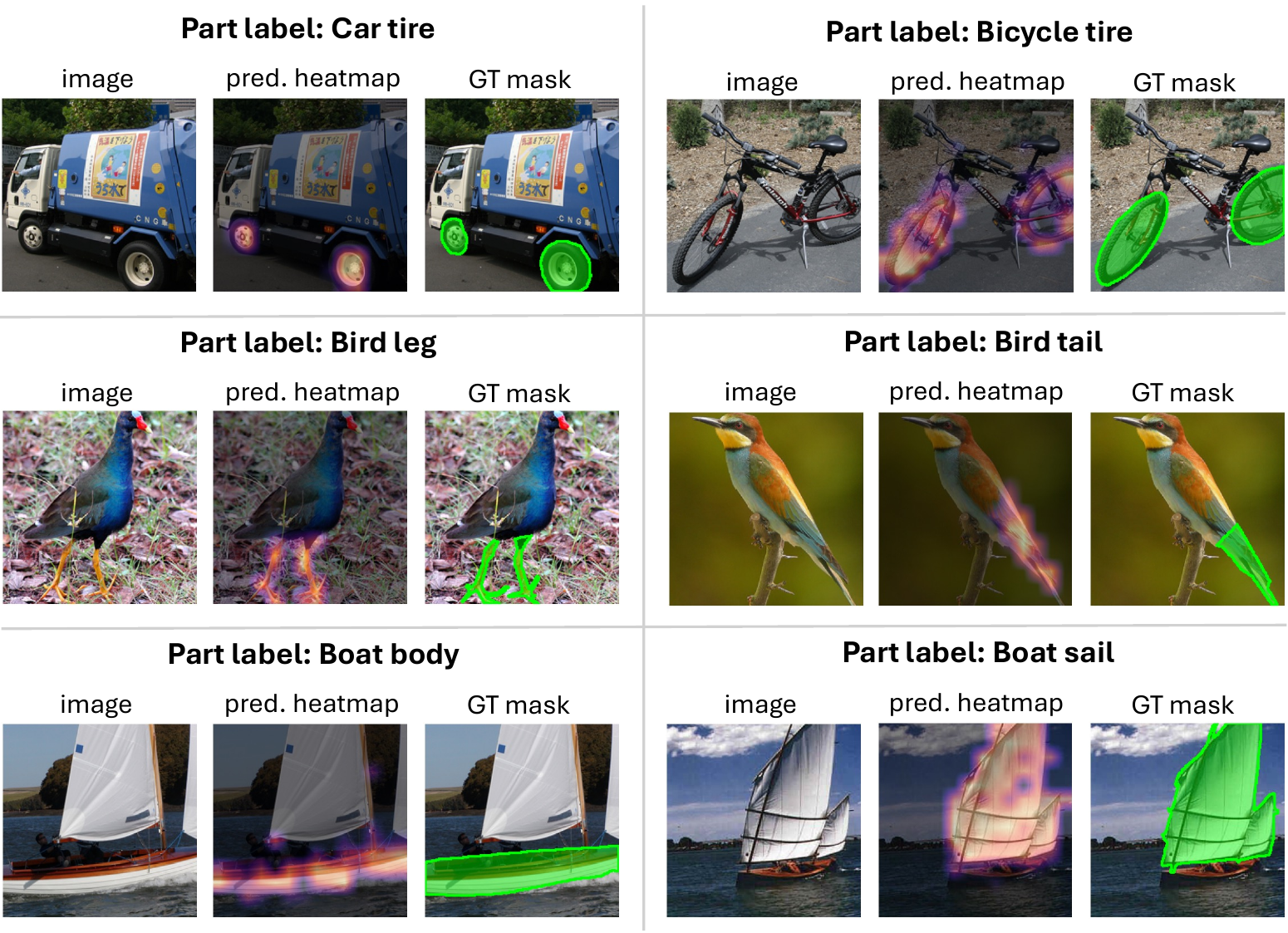}
\caption{Qualitative concept grounding results.}
\label{fig:qual_seg_results}
\end{figure}

We extend the results of the spatial concept grounding experiment. 
In Figure~\ref{fig:qual_seg_results} we show qualitative spatial grounding results of specific parts from the Part-ImageNet dataset, compared to their ground-truth segmentation masks. These results provide further qualitative evidence that our method provides high-quality localization of fine-grained concepts. 

\subsection{Local--Global compositional structure}

\begin{figure}
    \centering

    \includegraphics[width=0.55\linewidth]{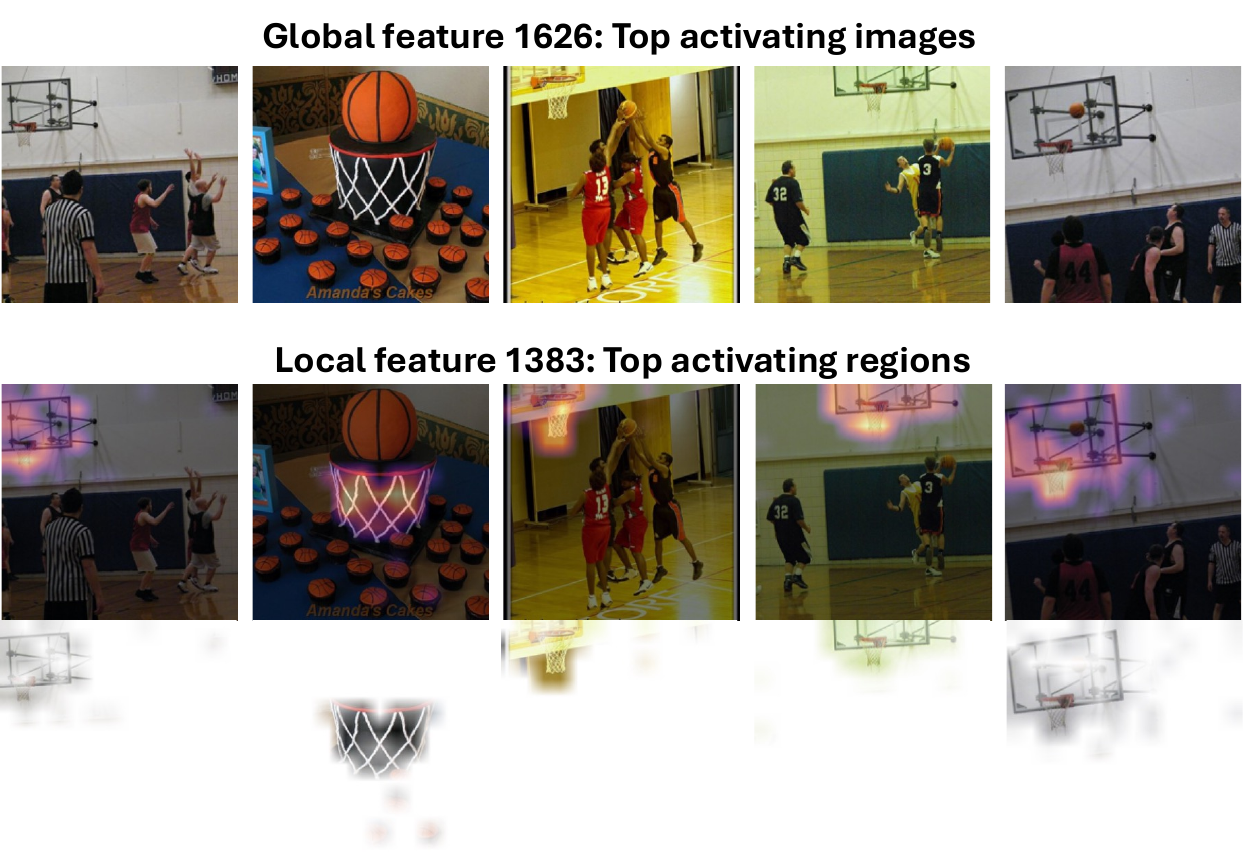}

    \vspace{0.5em}

    \includegraphics[width=0.55\linewidth]{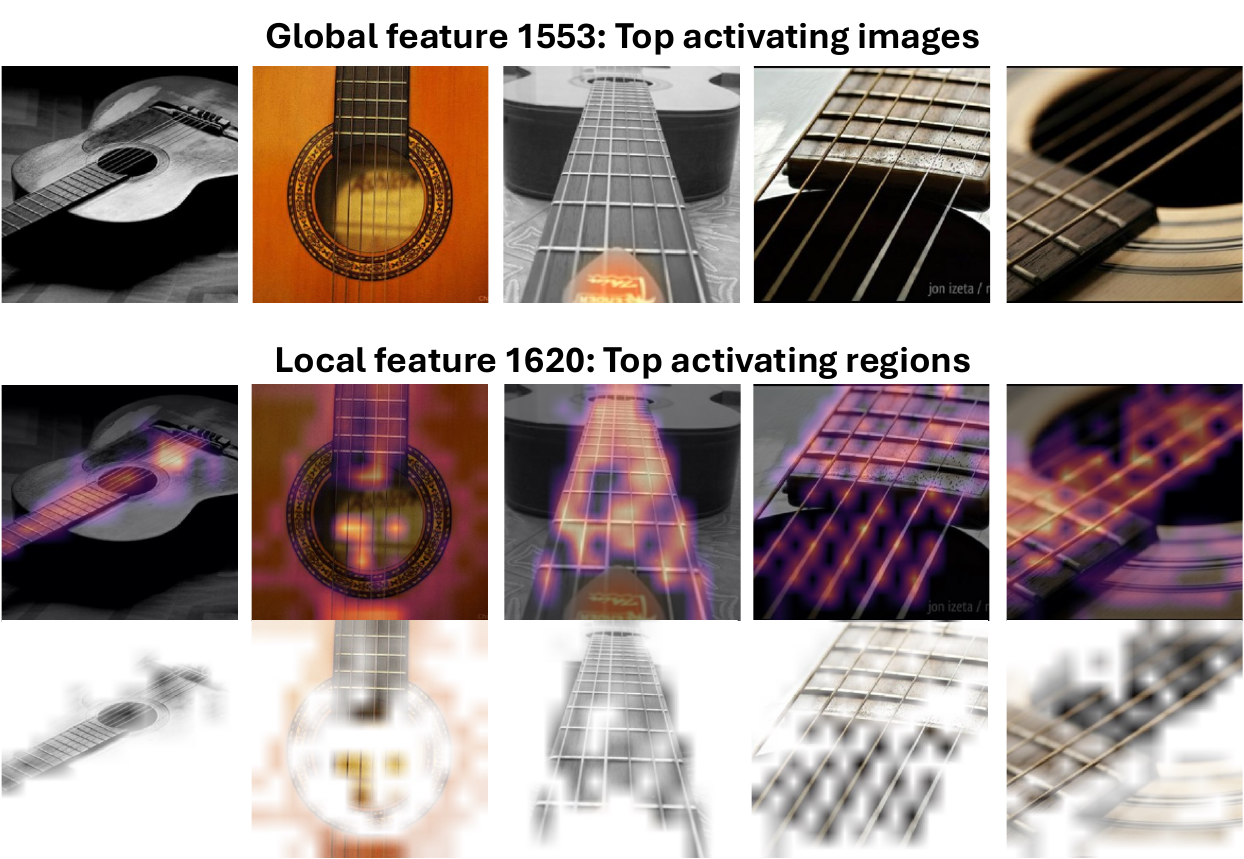}

    \vspace{0.5em}

    \includegraphics[width=0.55\linewidth]{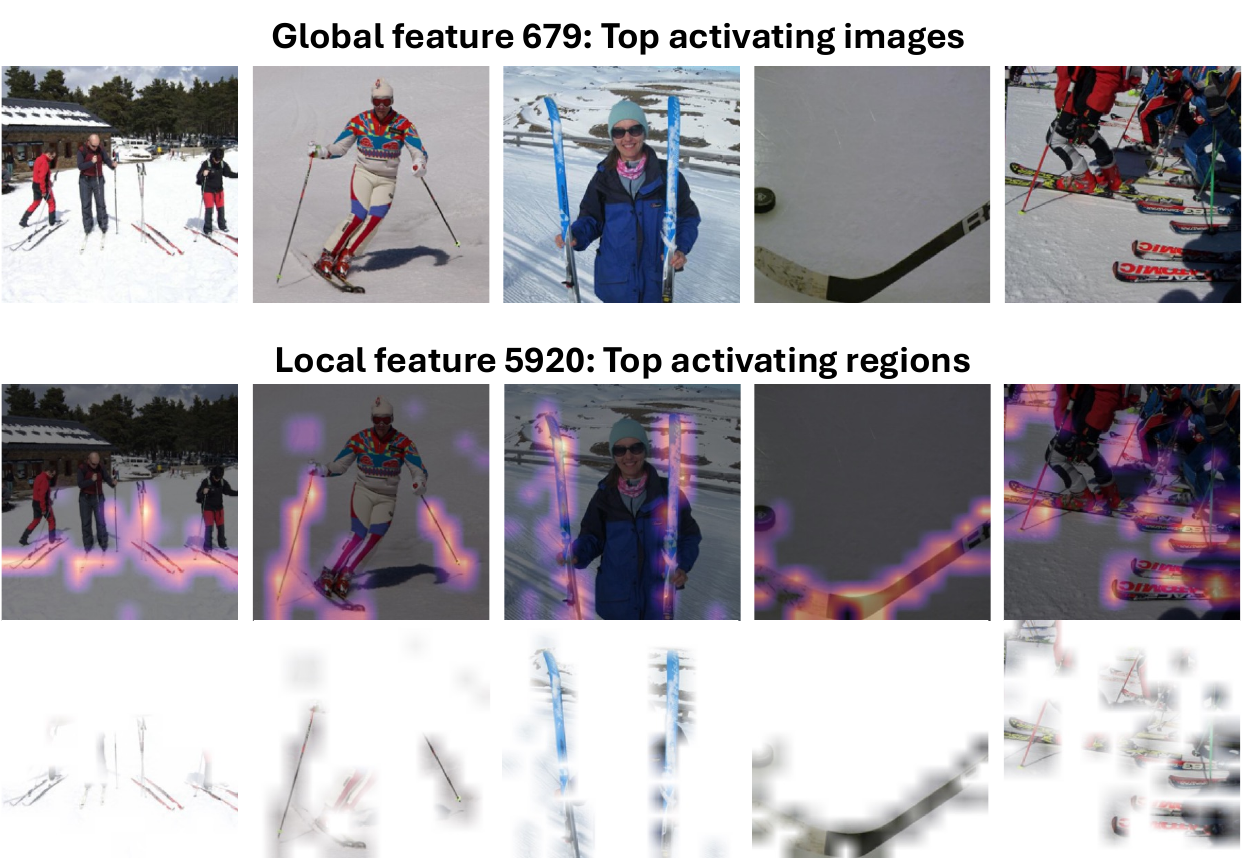}

    \vspace{0.5em}

    \includegraphics[width=0.55\linewidth]{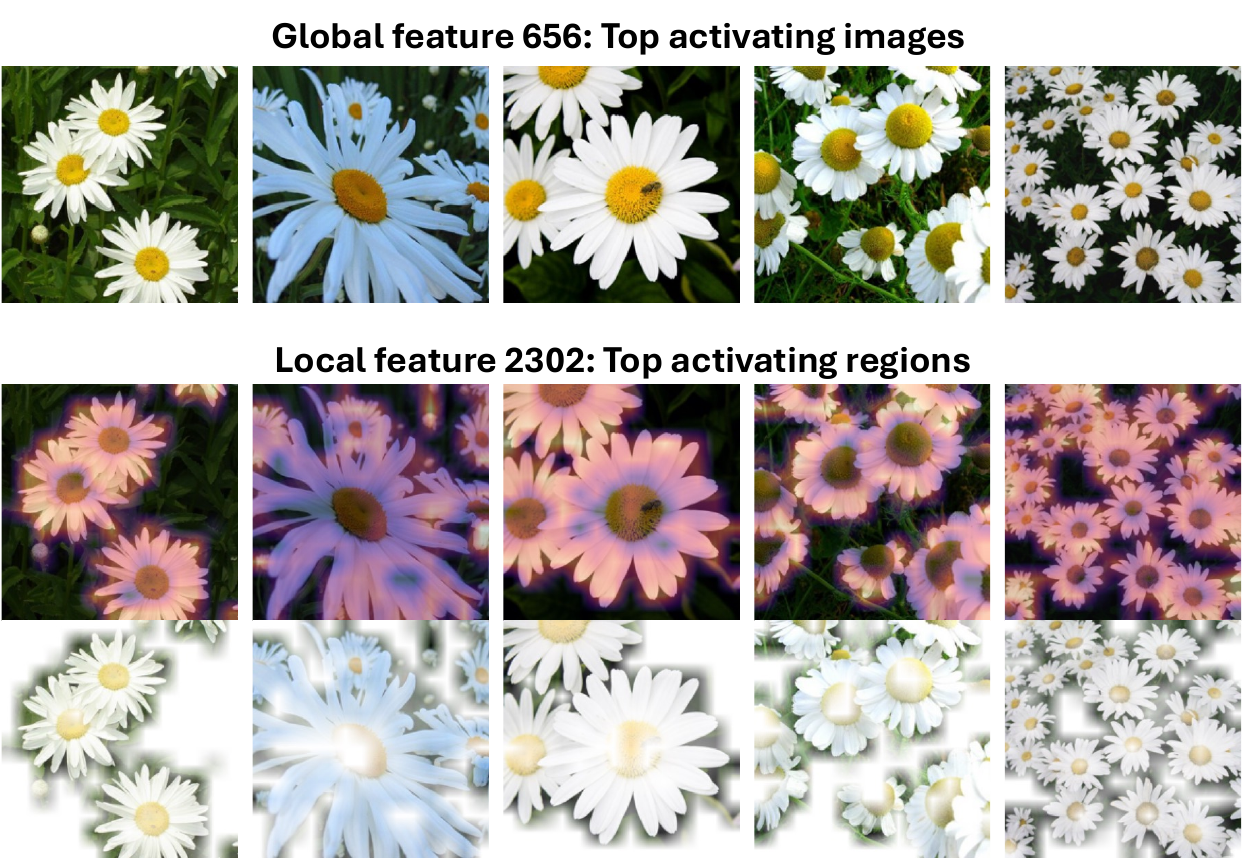}

    \caption{Qualitative results for LG-SAE local--global concept composition.}
    \label{fig:comp_results}
\end{figure}

We present additional qualitative examples demonstrating the meaningful local--global relationships captured by LG-SAE’s learned bridge.
We selected random entries from our learned bridge $\mathcal{C}_{i,j}$, connecting global SAE feature $i$ with local SAE feature $j$. In Fig.~\ref{fig:comp_results}, we show the five ImageNet validation images that most strongly activate the global feature $i$, and the corresponding spatial activation maps of its linked local feature $j$.

% \begin{figure}
%     \centering
%     \includegraphics[width=0.75\linewidth]{figs/comp_result_4.pdf}
%     % \caption{User instructions for concept composition survey.}
%     \label{fig:comp_result_4}
% \end{figure}

% \subsection{Spatially-Aware Concept Naming}
% ADD.

\subsection{Multiscale Explainability} 
\begin{figure}[H]
    \centering
    \includegraphics[width=0.9\linewidth]{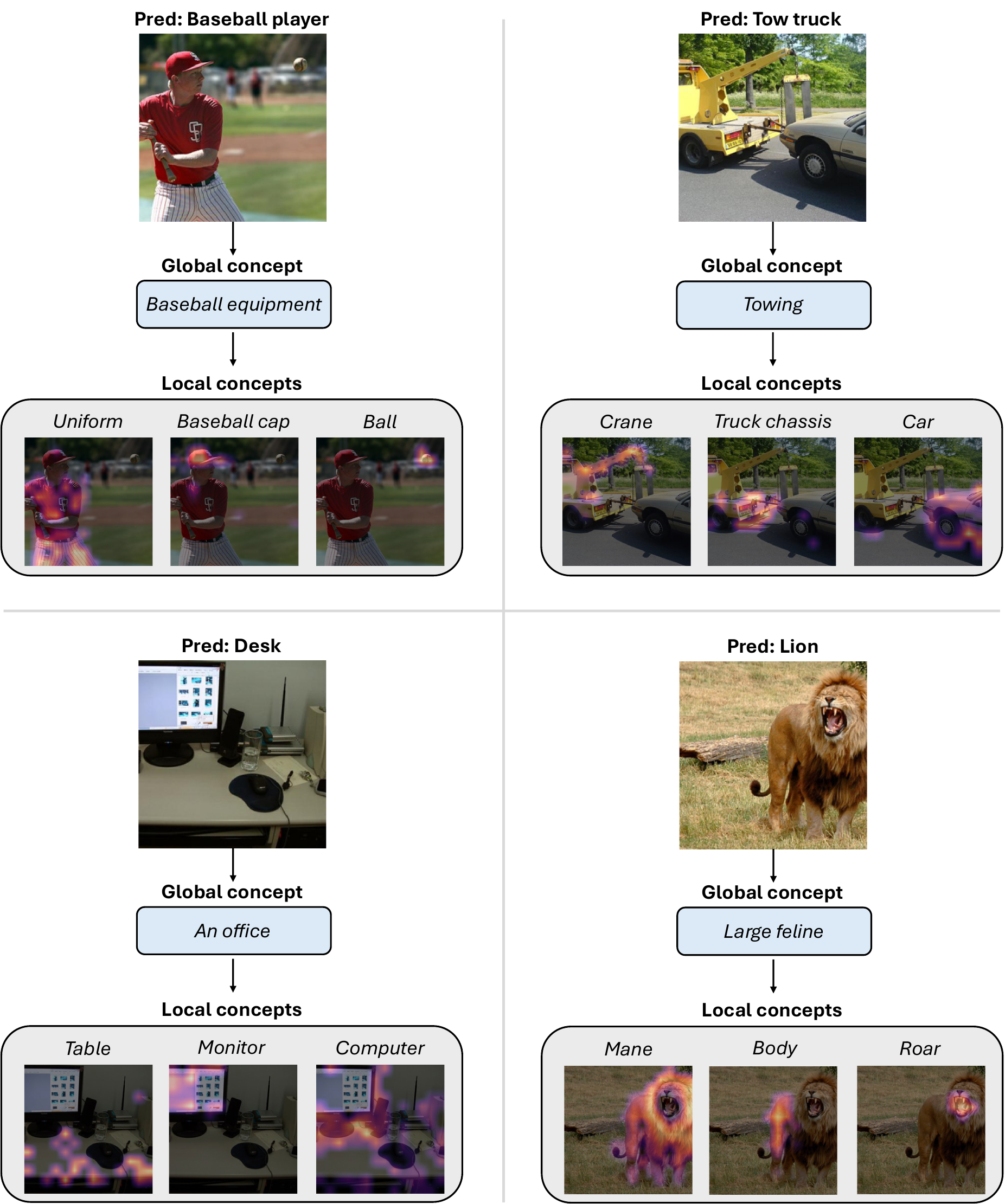}
    \caption{Multiscale model explainability examples. For each image, we show the predicted class, the global concept contributing most strongly to the prediction, and the three local concepts most related to it (i.e., with the highest relevance scores). We use CLIP-Dissect for naming global concepts, and our proposed spatially-aware method for naming local concepts.}
    \label{fig:qual_explanations}
\end{figure}

We demonstrate the multiscale explainability of LG-CBM, described in Section~\ref{LG-CBMs}.
In Figure~\ref{fig:qual_explanations} we show how predictions on ImageNet validation images are explained by global semantic concepts, along with their local constituents. For each image we present the global concept that contributed most strongly to the model's prediction, and the top three local concepts attached to it.

\clearpage

\end{document}